\documentclass[runningheads]{llncs}

\usepackage{graphicx}

\usepackage{tikz}
\usepackage{comment}

\usepackage{amsmath,amsfonts,bm}

\def\eqref#1{equation~\ref{#1}}

\def\1{\bm{1}}

\newcommand{\test}{\mathcal{D_{\mathrm{test}}}}

\def\ry{{\textnormal{y}}}

\def\va{{\bm{a}}}

\def\vg{{\bm{g}}}

\def\vm{{\bm{m}}}
\def\vn{{\bm{n}}}

\def\vs{{\bm{s}}}

\def\vw{{\bm{w}}}

\def\valpha{{\bm{\alpha}}}

\def\mF{{\bm{F}}}

\DeclareMathAlphabet{\mathsfit}{\encodingdefault}{\sfdefault}{m}{sl}
\SetMathAlphabet{\mathsfit}{bold}{\encodingdefault}{\sfdefault}{bx}{n}
\newcommand{\tens}[1]{\bm{\mathsfit{#1}}}
\def\tA{{\tens{A}}}
\def\tB{{\tens{B}}}

\def\tD{{\tens{D}}}

\def\tF{{\tens{F}}}

\def\tK{{\tens{K}}}

\def\tX{{\tens{X}}}
\def\tY{{\tens{Y}}}

\newcommand{\E}{\mathbb{E}}

\newcommand{\Var}{\mathrm{Var}}

\usepackage{amsmath,amssymb} 
\usepackage{color}
\usepackage{multirow}
\usepackage{booktabs}
\usepackage{enumitem}

\usepackage{listings}
\usepackage{xcolor}

\definecolor{codegreen}{rgb}{0,0.6,0}
\definecolor{codegray}{rgb}{0.5,0.5,0.5}
\definecolor{codepurple}{rgb}{0.58,0,0.82}
\definecolor{backcolour}{rgb}{0.95,0.95,0.92}

\lstdefinestyle{mystyle}{
	commentstyle=\color{codegreen},
	keywordstyle=\color{magenta},
	numberstyle=\tiny\color{codegray},
	stringstyle=\color{codepurple},
	basicstyle=\ttfamily\footnotesize,
	breakatwhitespace=false,         
	breaklines=true,                 
	captionpos=b,                    
	keepspaces=true,                 
	showspaces=false,                
	showstringspaces=false,
	showtabs=false,                  
	tabsize=2
}

\usepackage{array}
\newcolumntype{?}{!{\vrule width 1pt}}

\usepackage{siunitx}

\usepackage[width=122mm,left=12mm,paperwidth=146mm,height=193mm,top=12mm,paperheight=217mm]{geometry}

\usepackage[pagebackref,breaklinks,colorlinks]{hyperref}
\usepackage[capitalize]{cleveref}
\crefname{section}{Sec.}{Secs.}
\Crefname{section}{Section}{Sections}
\Crefname{table}{Table}{Tables}
\crefname{table}{Tab.}{Tabs.}

\usepackage{siunitx}
\usepackage{booktabs}
\usepackage{etoolbox}
\renewrobustcmd{\bfseries}{\fontseries{b}\selectfont}
\renewrobustcmd{\boldmath}{}
\newrobustcmd{\B}{\bfseries}

\usepackage{stackengine}
\usepackage{caption}
\usepackage{subcaption}
\usepackage{pifont}%
\DeclareMathOperator{\fc}{fc}
\DeclareMathOperator{\APN}{APN}

\DeclareMathOperator{\Area}{Area}

\newcommand{\norm}[1]{\| #1 \|}

\newcommand{\etal}{\textit{et al.}}

\usepackage[normalem]{ulem}

\begin{document}
	\pagestyle{headings}
	\mainmatter
	\def\ECCVSubNumber{4222}  
	
	\title{Image Inpainting with Cascaded Modulation GAN and Object-Aware Training}

	\titlerunning{CM-GAN: Image Inpainting with CM-GAN and Object-Aware Training}
	\vspace{-7mm}
	\small{\author{Haitian~Zheng\textsuperscript{1,2} \and Zhe~Lin\textsuperscript{2} \and Jingwan~Lu\textsuperscript{2} \and Scott~Cohen\textsuperscript{2} \and Eli~Shechtman\textsuperscript{2} \and Connelly~Barnes\textsuperscript{2} \and Jianming~Zhang\textsuperscript{2} \and Ning~Xu\textsuperscript{2} \and Sohrab~Amirghodsi\textsuperscript{2} \and Jiebo~Luo\textsuperscript{1}}
	}
	\authorrunning{H. Zheng, Z. Lin et al.}
	%
	\vspace{-3mm}
	\institute{University of Rochester \and Adobe Research}
	\maketitle

	\vspace{-5mm}
	\begin{figure*}[h]
		\centering

		\includegraphics[width=1\linewidth]{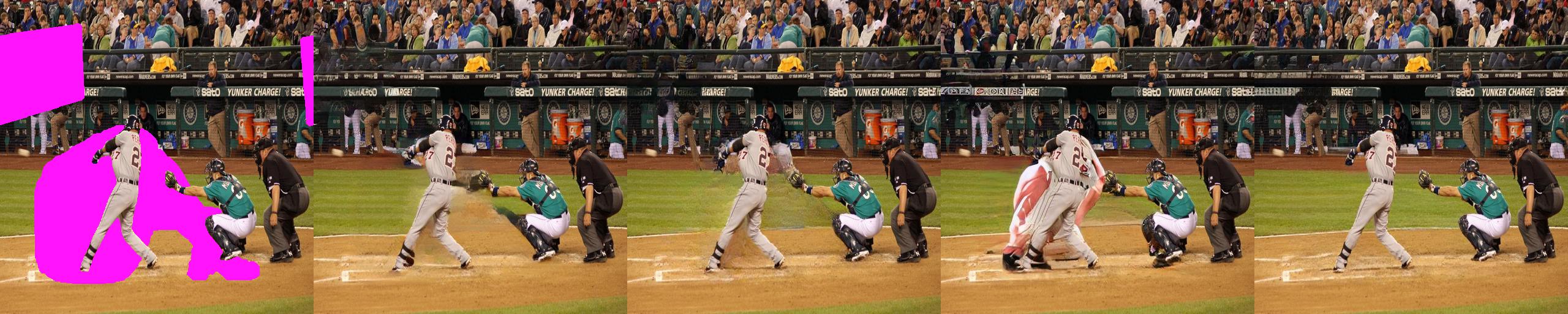}
		\includegraphics[width=1\linewidth]{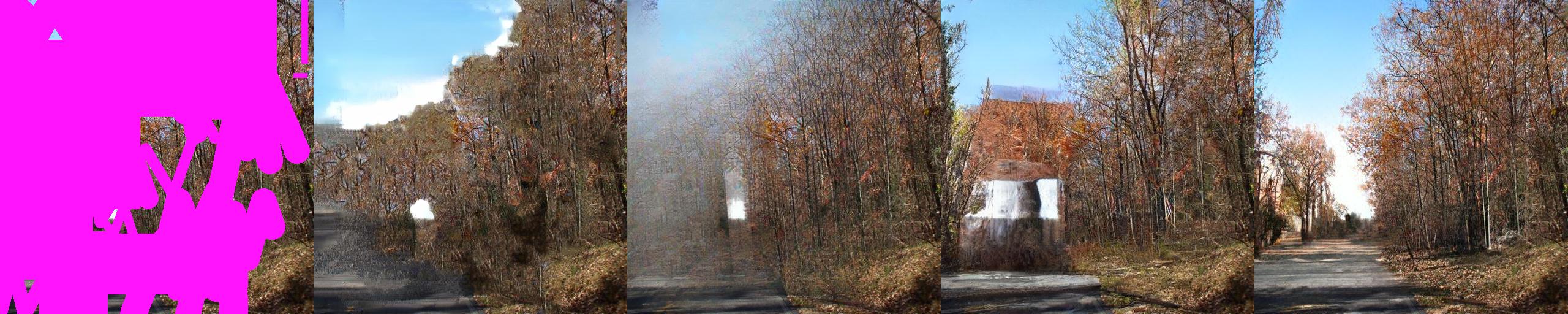}

		\includegraphics[width=1\linewidth]{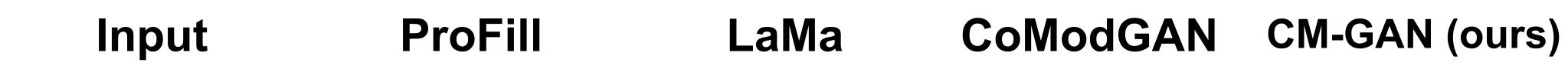}
		\vspace{-0.1in}
		\caption{
			\small
			Results of CM-GAN in comparison to state of the art methods: ProFill~\cite{profill},
			LaMa~\cite{lama} and
			CoModGAN~\cite{comodgan}.
			CM-GAN generates more plausible and realistic results for the distractor removal scenario (1st row) and large holes (2nd row).
			\vspace{-0.4in}
		}
		\label{fig:teaser}
	\end{figure*}

	\vspace{-1mm}
	\begin{abstract}
		Recent image inpainting methods have made great progress but often struggle to generate plausible image structures when dealing with large holes in complex images. This is partially due to the lack of effective network structures that can capture both the long-range dependency and high-level semantics of an image. We propose cascaded modulation GAN (CM-GAN), a new network design consisting of an encoder with Fourier convolution blocks that extract multi-scale feature representations from the input image with holes and a dual-stream decoder with a novel cascaded global-spatial modulation block at each scale level. In each decoder block, global modulation is first applied to perform coarse and semantic-aware structure synthesis, followed by spatial modulation to further adjust the feature map in a spatially adaptive fashion. In addition, we design an object-aware training scheme to prevent the network from hallucinating new objects inside holes, fulfilling the needs of object removal tasks in real-world scenarios. Extensive experiments are conducted to show that our method significantly outperforms existing methods in both quantitative and qualitative evaluation.
		Please refer to the project page: \url{https://github.com/htzheng/CM-GAN-Inpainting}.
		\keywords{Image Inpainting, Generative Adversarial Networks}
	\end{abstract}

	\section{Introduction}

Image inpainting refers to the task of completing missing regions of an image as shown in \cref{fig:teaser}. It is one of the fundamental tasks in computer vision and has many practical applications, such as object removal~\cite{profill,deepfillv2} and manipulation~\cite{sesame,oh2001image}, image re-targeting~\cite{setlur2005automatic,vaquero2010survey,patchmatch}, image compositing~\cite{chen2019toward}, and 3D photo effects~\cite{Niklaus_TOG_2019,Kopf-OneShot-2020}.

Early inpainting methods leverage patch-based synthesis~\cite{patchmatch,kwatra2005texture,cho2008patch} or color diffusion~\cite{ballester2001filling,chan2001nontexture,shen2002mathematical,criminisi2004region} to fill the holes by propagating repeating textures and patterns from the visible regions.
To facilitate the completion of more complex image structures, recent research efforts have shifted to adopting a data-driven scheme where deep generative networks are learned to predict visual content and appearance~\cite{deepfill,deepfillv2,profill,lama,comodgan}.
By training on a large corpus of images and with assist of both reconstruction and adversarial losses, generative inpainting models have shown to produce more visually appealing results on various types of input data including natural images and faces.



While existing works have shown promising results on completing simple image structures, generating complex holistic structures and image contents with high-fidelity details remains a huge challenge, espeically when the holes are large.
Essentially, how to 1) \emph{accurately propagate global context into the incomplete region} while 2) \emph{synthesizing realistic local details that are coherent to the global clue} is the key question for image inpainting.
To tackle global context propagation, existing networks leverage encoder-decoder structure~\cite{pathak2016context,iizuka2017globally,park2020swapping}, dilated convolution~\cite{yu2015multi,deepfillv2}, contextual attention~\cite{deepfill,deepfillv2,hifill},
or Fourier convolution~\cite{lama} to incorporate long range feature dependency for expanding the effective receptive field~\cite{luo2016understanding}.
Furthermore, two-stage approaches~\cite{YangLLSWL17,deepfill,deepfillv2,edgeconnect,foreground_aware} and iterative hole filling~\cite{profill} predict a coarse result such as a smoothed image, edge/semantic maps or partial completion to enhance the global structure. 
However, those models lack a mechanism to capture high-level semantics in the unmasked region and effectively propagate them into the hole to synthesize a holistic global structure.
With the typical shallow bottleneck designs, the designed feature propagation layer is less aware of global semantics and more prone to generating incoherent local details potentially leading to visual artifacts.

More recently, feature modulation-based methods~\cite{stylegan,stylegan2,park2020swapping} have shown very promising results on controlling image generation with a global style code.
Benefiting from a global code that captures the context of the entire image, CoModGAN~\cite{comodgan} attempts to inject global context into the generator for filling in very large holes~\cite{comodgan}.
However, due to the lack of spatial adaptation, global modulation is sensitive to the corrupted encoding feature inside the inpainting region (shown in \cref{fig:feature}). Their results show that passing global information in this way is insufficient for synthesizing high quality global structures which may lead to severe artifacts such as the large unseen color blobs~\cite{comodgan} or inconsistent visual appearances such as distorted structures,
cf. \cref{fig:main_compare}.

To seek a better way to inject global context into the missing region in inpainting, we investigate a new modulation scheme by cascading global and spatial modulations. We propose \textbf{C}ascaded \textbf{M}odulation \textbf{GAN} (\textbf{CM-GAN}), a new generative network that can synthesize better holistic structure and local details, cf. \cref{fig:teaser} and \cref{fig:main_compare}. 
Different from~\cite{comodgan} that attempts to globally modulates the partially invalid encoder feature as shown \cref{fig:feature}, our  
spatially-adaptive modulation scheme cascaded after a global modulation is much more effective in processing invalid features inside the hole.
Although several spatially-adaptive modulation schemes~\cite{spade,stylemapgan} have been proposed in the past to tackle inpainting, our cascaded modulation approach is significantly different from those works in that: i) our spatial code comes from the \emph{decoding stage} rather than the encoding stage to avoid modulating the decoder with invalid encoding feature (cf.~\cref{fig:feature}), 2) we incorporate the global code into spatial modulation for enforcing the global-local consistency, 3) we introduce a dual-branch design that decouples global and local features for structure-details separation, and 4) we design spatially-aware demodulation instead of instance or batch normalization to avoid potential `droplet artifact''~\cite{stylegan2}.
Furthermore, on the encoder side, we inject the Fast Fourier convolution~\cite{ffc} at each stage of the encoder network to expand the receptive field of the encoder at early stages, allowing the network to capture long-range correlations across the image. 


Another design aspect to consider is how to generate synthetic masks used during inpainting training. We need to design the masks tailored for real-world inpainting use cases, such as object removal and partial object completion. Previous methods generate training data by randomly locating rectangular ~\cite{pathak2016context,iizuka2017globally} or irregularly-shaped~\cite{deepfillv2,comodgan} masks.
However, the masks users draw in common use cases likely have a shape of an existing object, or part of an object or other relatively simple shapes such as a scribble or a blob. Moreover, users usually expect the removed objects to be filled by background textures and structures, thus new objects are not expected to appear inside the holes.  
However, models trained with randomly located masks tend to have the color-bleeding effect across object boundaries and generate object-like artifact blobs inside the hole~\cite{comodgan}.
In a better attempt, a recent work~\cite{profill} leverages saliency annotation to simulate the holes left by distracting objects occluded by the foreground salient objects.
However, saliency annotations only capture large dominant foreground objects, thus resulting in the algorithm constructing large holes that include most of the other less salient objects. This is different from the real use cases where only a few distracting objects need to be removed.

We found that the tendency to generate spurious objects, blobs and color-bleeding effects across object boundaries can be addressed by a proper and more carefully designed object-aware training scheme.
In terms of mask sampling for training, different from random mask~\cite{deepfillv2,comodgan,lama,partial_conv,structureflow} or saliency-based mask~\cite{profill}, we leverage an instance-level panoptic segmentation model~\cite{panopticFCN} to generate object-aware masks that better simulate real distractors or clutter removal use cases.
To avoid generating distorted objects or color blobs inside the hole, we filter out cases where the entire object or a large part of the object is covered by the mask. Furthermore, the panoptic segmentation provides precise object boundaries and thus prevents the trained model from leaking colors at object boundaries.

Finally, for the training losses, we propose a masked $R_1$ regularization specifically designed for inpainting and augment the adversarial loss with a perceptual loss extracted by segmentation model for improving robustness. 
The new regularization avoids penalizing the model outside the mask and thus imposes a better separation of input condition from the generated contents.
Consequently, the new regularization eliminates the potential harmful impact of computing regularization on the background.
Our contributions are four-fold:
\begin{itemize}[noitemsep]
    \item Cascaded Modulation GAN, a new inpainting network architecture formed by a masked image encoder with Fourier convolution blocks and a cascaded global-spatial modulation-based decoder.
    
    \item An object-aware mask generation scheme preventing the model from generating new objects inside holes and mimicking realistic inpainting use cases.
    
    \item A masked $R_1$ regularization loss to stabilize the adversarial training for the inpainting tasks.
    
    \item State-of-the-art results on the Places2 dataset for various types of masks.
\end{itemize}


	\section{Related works}
\subsection{Image Inpainting}
Image inpainting has a long standing history.
Traditionally, the patch-based methods~\cite{Image_quilting,kwatra2005texture,patchmatch,cho2008patch} search and copy-paste patches from known region to progressively fill in the target hole. Meanwhile, the diffusion-based methods~\cite{ballester2001filling,chan2001nontexture,shen2002mathematical,criminisi2004region} describe and solve the color propagation inside the hole via partial differential equation.
The above methods can produce high-quality stationary textures while completing simple shapes, but they lack the mechanisms to model the high-level semantics for completing new semantic structure inside the hole.
Recently, inpainting methods have shifted to a data-driven scheme where deep generative models are learned to directly predict the filled in content inside the hole in an end-to-end fashion.
By training deep generative models via adversarial training~\cite{gan}, the learned model can capture higher-level representation of images and hence can generate more visually plausible results.
Specifically, Pathak \etal~\cite{pathak2016context} first leverage an encoder-decoder network with a bottleneck layer to predict the missing structures for hole filling.
Iizuka \etal~\cite{iizuka2017globally} propose a two discriminator design to encourage the global and local consistency separately.
To make the generator better at capturing the global context, several attempts have been made.
Motivated by the structure-texture decomposition principle~\cite{aujol2006structure,bertalmio2003simultaneous}, two-stage networks predict an intermediate representation of image with smoothed image~\cite{deepfill,MEDFE,hifill,ict,profill}, edge~\cite{edgeconnect,foreground_aware}, gradient~\cite{yang2020learning} or segmentation map~\cite{song2018spg} for enhancing the final output.
Yu \etal~\cite{deepfill} design contextual attention to explicitly let the network borrow patch features at a global scale.
Aiming at expanding the receptive field of the network, Iizuka \etal~\cite{iizuka2017globally} Yu \etal~\cite{deepfill} incorporate dilated convolutions to the generator. Likewise, Suvorov \etal~\cite{lama} leverage Fourier convolution~\cite{ffc} to acquire a global receptive field.
Furthermore, feature gating such as partial convolution~\cite{partial_conv} and gated convolution~\cite{deepfillv2} is proposed to handle invalid features inside the hole.
To enhance global prediction capacity, Zhao \etal~\cite{comodgan} propose an encoder-decoder network that leverages style code modulation for global-level structure inpainting. 
To augment the adversarial loss and to suppress artifacts, existing works~\cite{deepfill,deepfillv2,lama,profill,edgeconnect} often train the generator with additional reconstruction objectives such as $\ell_1$, perceptual~\cite{johnson2016perceptual} or contextual~\cite{pathak2016context} loss. Recently, Suvorov \etal~\cite{lama} propose to use segmentation networks to compute perceptual loss which achieve better performance.

\subsection{Feature Modulation for Image Synthesis}
Originating from style transfer~\cite{huang2017arbitrary}, feature modulation~\cite{huang2018multimodal,stylegan,stylegan2} has been widely adopted to incorporate input conditions for controlled generation~\cite{wang2018recovering,park2020swapping,albahar2021pose,comodgan,spade,zheng2020example,tan2020semantic}.
Existing modulation methods usually leverage batch normalization or instance normalization to normalize the input feature. The modulation is then achieved by scaling and shifting the normalized activation according to affine parameters predicted from input conditions.
Recently, Karras \etal~\cite{stylegan2} find that normalization would cause the ``droplet artifact'' as the network can create a strong activation spike to sneak signal through normalization layer.
Consequently, StyleGAN2~\cite{stylegan2} replaces the feature normalization with a proposed demodulation step~\cite{stylegan2} for better image synthesis.
To modulate the input feature according to a spatially-varying feature map, spatial modulation~\cite{spade,stylemapgan,albahar2021pose,wang2018recovering} are proposed. Essentially, those methods leverage convolutional layers to predict 2d affine parameters for spatially-controlled modulation. However, feature normalization makes the existing approaches~\cite{albahar2021pose,wang2018recovering} less consistent with the design principle of StyleGAN2.

\subsection{Regularization for Adversarial Training}
Adversarial training is known to be  challenging~\cite{mescheder2018training} as it is hard for the adversarial networks to reach global Nash equilibrium~\cite{fid}.
Consequently, various regularization are proposed to stabilize the GAN training. In particular, weight normalization~\cite{salimans2016weight} and spectrum normalization~\cite{miyato2018spectral} are proposed to constrain the the Lipschitz continuity of the discriminator. Likewise, Gulrajani \etal~\cite{gulrajani2017improved} propose a gradient penalty to impose a $K$-Lipschitz constraint to the discriminator.
Mescheder \etal~\cite{mescheder2018training} propose $R_1$ regularization to penalize the discriminator gradient on real data, which is later used by~\cite{stylegan2,comodgan,lama}.
Karras \etal~\cite{stylegan2} propose perceptual path length regularization on the generator to ensure smoothness mappings and lazy regularization to optimize the training efficiency.

\section{Method}\label{sec:methodology}

\subsection{Cascaded Modulation GAN}\label{sec:cmgan}
To better model the global context for image completion~\cite{deepfill,deepfillv2,profill,crfill,hifill,comodgan,lama}, we propose a novel mechanism that \emph{cascades global code modulation with spatial code modulation} to facilitate the processing of the partially invalid feature while better injecting the global context into the spatial region. It leads to a new architecture named Cascaded Modulation GAN (CM-GAN), which can synthesize holistic structures and local details surprisingly well as shown in \cref{fig:teaser}.

\begin{figure}[t]
	\centering
    \includegraphics[width=1.0\linewidth]{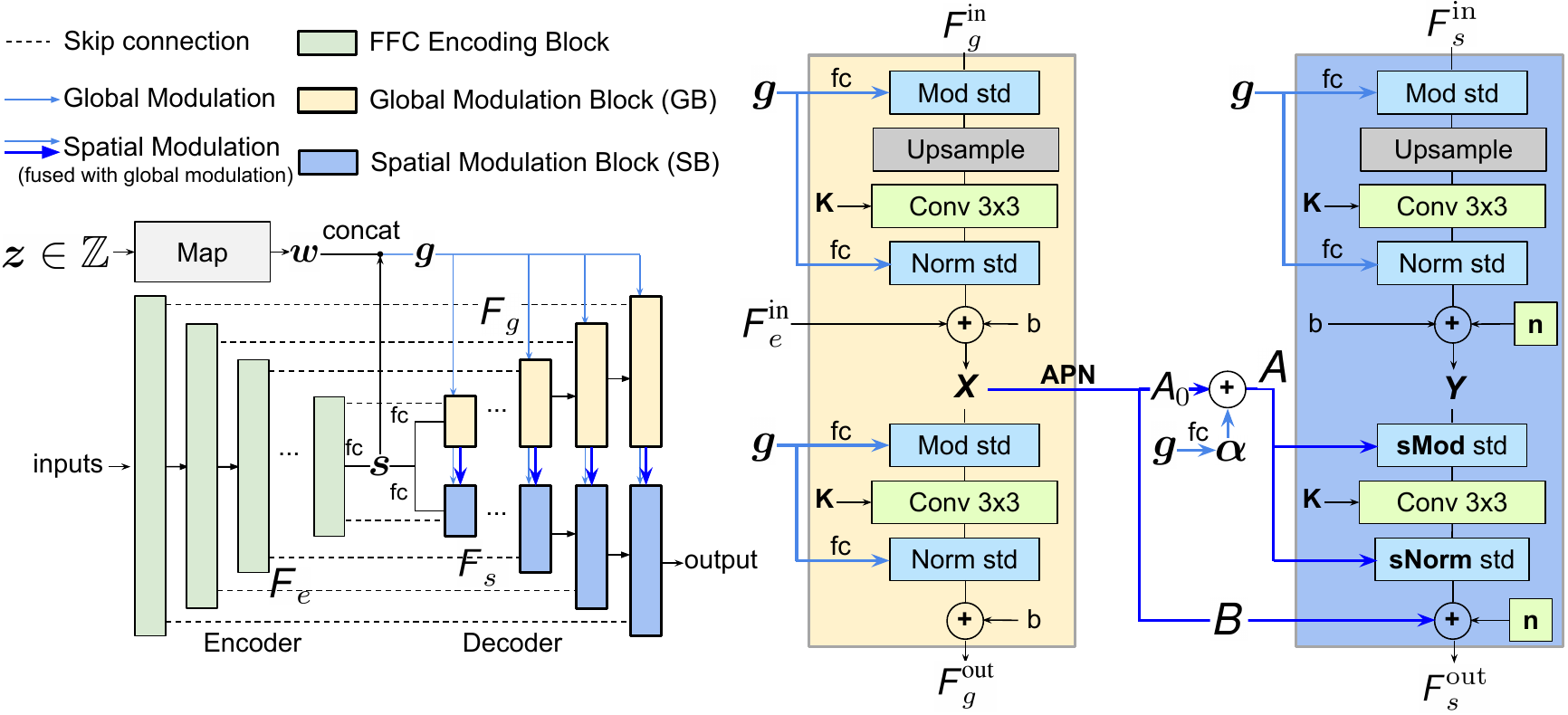}
	\caption{
    \small
    \textbf{Left}: The CM-GAN architecture, which consists of an encoder with FFC blocks and a two-stream decoder with a cascade of global modulation block (GB) and subsequent spatial modulation block (SB).
    This cascaded modulation scheme extracts spatial style codes from the globally modulated feature map (instead of from the encoder feature map used in the previous work) to make spatial modulation more effective for inpainting. \textbf{Right}: Cascaded modulation at each scale.
    GB and SB take $\tF_g^{\text{in}}$ and $\tF_s^{\text{in}}$, respectively, as inputs and produce the upsampled feature $\tF_g^{\text{out}}$ and $\tF_s^{\text{out}}$.
    Specifically, we apply joint global-spatial modulation to ensure the generation consistency both at the global and local scales.
	}
	\label{fig:network}
	\vspace*{-.15in}
\end{figure}

\noindent \textbf{Network Overview.} \quad 
As shown in \cref{fig:network} (left), CM-GAN is based on an encoder branch and two parallel cascaded decoder branches to generate visual output.
Specifically, it starts with an encoder that takes the partial image and the mask as inputs to produce multi-scale feature maps $\tF_e^{(1)},\cdots,\tF_e^{(L)}$ at each scale $1 \le i \le L$ ($L$ is the highest level with the smallest spatial size). Unlike most encoder-decoder methods~\cite{deepfill,lama} and to facilitate the completion of holistic structure, we extract a global style code $\vs$ from the highest level feature $\mF_e^{(L)}$ with a fully connected layer followed by a $\ell_2$ normalization.
Furthermore, an MLP-based mapping network~\cite{stylegan2} is used to generate a style code $\vw$ from noise, simulating the stochasticity of image generation. The code $\vw$ is joined with $\vs$ to produce a global code $\vg=[\vs;\,\vw]$ for the consequent decoding steps.

\noindent \textbf{Global-Spatial Cascaded Modulation.} \quad 
To better bridge the global context at the decoding stage, we propose \emph{global-spatial \textbf{C}ascaded \textbf{M}odulation} (CM).
As shown in \cref{fig:network} (right), the decoding stage is based on two branches of Global Modulation Block (GB) and Spatial Modulation Block (SB) to respectively upsample global feature $\tF_g$ and local features $\tF_s$ in parallel.
Different from existing approaches~\cite{deepfillv2,profill,ict,lama,comodgan}, the CM design introduces a new way to inject the global context into the hole region.
At a conceptual level, it consists of a cascade of global and spatial modulations between features at each scale and naturally integrates three compensating mechanisms for global context modeling: 1) \emph{feature upsampling} allows both GB and SB to utilize the global context from the low-resolution features generated by both of the previous blocks; 2) the \emph{global modulation} (cyan arrows of \cref{fig:network}) allows both GB and SB to leverage the global code $\vg$ for generating better global structure; and 3) \emph{spatial modulation} (blue arrows of \cref{fig:network}) leverages spatial code (intermediate feature output of GB) to further inject fine-grained visual details to SB.

More specifically, as shown in \cref{fig:network} (right), CM at each level of the decoder consists of the paralleled GB block (yellow) and SB block (blue) bridged by spatial modulation. Such parallel blocks takes $\tF_g^{\text{in}}$ and $\tF_s^{\text{in}}$ as input and output $\tF_g^{\text{out}}$ and $\tF_s^{\text{out}}$.
In particular, GB leverages an initial upsampling layer following a convolution layer to generate the intermediate feature $\tX$ and global output $\tF_g^{\text{out}}$, respectively. Both layers are modulated by the global code $\vg$~\cite{stylegan2} to capture the global context.

\begin{figure}[t]
	\centering
	\includegraphics[width=1.\linewidth]{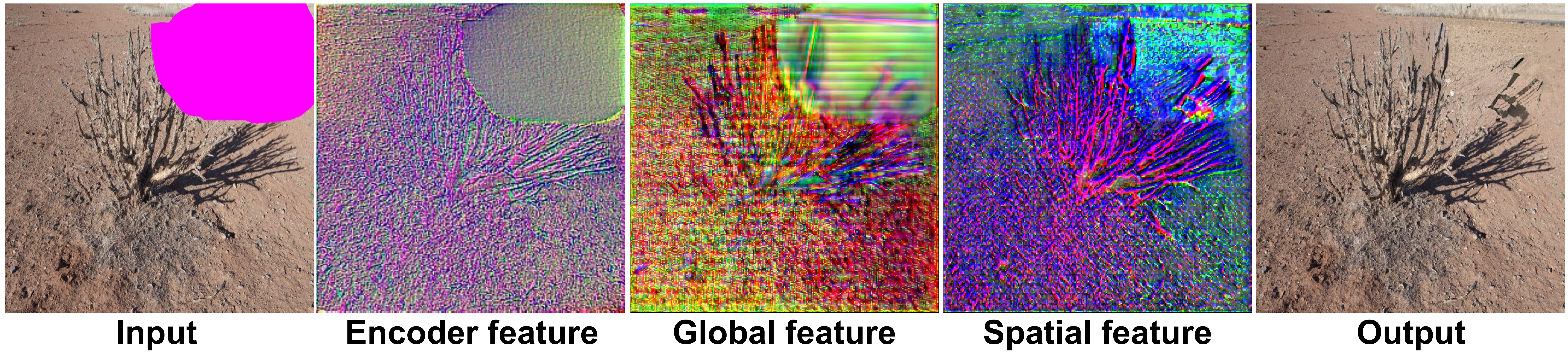}
	\caption{
	\small
    Visualization of the intermediate features for inpainting. From left to right are the incomplete image, encoded feature, our globally modulated feature and spatially modulated features at the $256\times256$ layer and the output image.
	}
	\label{fig:feature}
	\vspace*{-.15in}
\end{figure}

Due to the limited expressive power of the global code $\vg$ to represent a $2$-d scene, and the noisy invalid features inside the inpainting hole~\cite{deepfillv2,partial_conv}, the global modulation alone generates distorted features inconsistent with the context as shown in \cref{fig:feature} and leads to visual artifacts such as large color blobs and incorrect structure as demonstrated in \cref{fig:ablation_CM}.
To address this critical issue, we cascade GB with an SB to correct invalid features while further injecting spatial details. SB also takes the global code $\vg$ to synthesize local details while respecting global context.
Specifically, taking the spatial feature $\tF_s^{\text{in}}$ as input, SB first produces an initial upsampled feature $\tY$ with an upsampling layer modulated by global code $\vg$. Next, $\tY$ is jointly modulated by $\tX$ and $\vg$ in a spatially adaptive fashion following the
\emph{modulation-convolution-demodulation} principle~\cite{stylegan2}:
\begin{itemize}
    \item \emph{Global-spatial feature modulation.}
    A spatial tensor $\tA_0 = \APN(\tX)$ is produced from feature $\tX$ by a 2-layer convolutional affine parameter network (APN). 
    Meanwhile, a global vector $\valpha = \fc(\vg)$ is produced from the global code $\vg$ with a fully connected layer ($\fc$) to incorporate the global context.
    Finally, a fused spatial tensor $\tA = \tA_0 + \valpha$ leverages both global and spatial information extracted from $\vg$ and $\tX$, respectively, to scale the intermediate feature $\tY$ with element-wise product $\odot$:
        \begin{align}
        \label{eq:modulation_scale2}
            \bar{\tY} &= \tY \odot \tA.
        \end{align}
    \item \emph{Convolution}. The modulated tensor $\bar{\tY}$ is then convolved with a $3\times3$ learnable kernel $\tK$, resulting in $\hat{\tY}$
        \begin{align}
        \label{eq:conv}
            \hat{\tY} &= \bar{\tY} \ast \tK.
        \end{align}
    \item \emph{Spatially-aware demodulation}. Different from existing spatial modulation methods~\cite{spade,stylemapgan,albahar2021pose}, we discard instance or batch normalization to avoid the known ``water droplet'' artifact~\cite{stylegan2} and propose a spatially-aware demodulation step to produce normalized output $\widetilde{\tY}$.
    Specifically, we assume that the input features $\tY$ are independent random variables with unit variance and after the modulation, the expected variance of the output does not change, i.e. $\displaystyle \E_{\ry \in \widetilde{\tY}} [\Var(\ry)]=1$.
    This assumption gives the demodulation computation:
        \vspace{-0.3cm}
        \begin{align}
        \label{eq:demodulation}
        \widetilde{\tY} &= \hat{\tY} \odot \tD,
        \end{align}
    where $\tD = 1 / \sqrt{\tK^2 \odot \E_{\va \in \tA}[\va^2]}$ is the demodulation coefficient. \cref{eq:demodulation} is implemented with standard tensor operations as elaborated in the supplementary material.
    \item \emph{Adding spatial bias and broadcast noise}.
    To introduce further spatial variation from feature $\tX$,
    the normalized feature $\widetilde{\tY}$ is added to a shifting tensor $\tB = \APN(\tX)$ produced by another affine parameter network from feature $\tX$ along with the broadcast noise $\vn$ to generate the new local feature $\tF_s^{\text{out}}$:
    \begin{align}
    \label{eq:modulation_shift}
        \tF_s^{\text{out}} = \widetilde{\tY} + \tB + \vn.
    \end{align}
\end{itemize}
As shown in the 4th column of \cref{fig:feature}, the cascaded SB block helps generate fine-grained visual details and improves the consistency of feature values inside and outside the hole.

\noindent \textbf{Expanding the Receptive Field at Early Stages.} \quad
The fully convolutional models suffer from slow growth of the effective receptive field~\cite{luo2016understanding}, especially at early stages of the network.
For this reason, an encoder based on strided convolution usually generates invalid features inside the hole region, making the feature correction at the decoding stage more challenging.
A recent work~\cite{lama} shows that fast Fourier convolution (FFC)~\cite{ffc} can help early layers achieve large receptive fields that cover the entire image.
However, the work~\cite{lama} stacks FFC at the bottleneck layer and is computationally demanding. Moreover, like many other works~\cite{deepfillv2}, due to the shallow bottleneck layers, ~\cite{lama} cannot capture global semantics effectively, limiting its ability to handle large holes.
We propose to replace every convolutional blocks of the CNN encoder with FFC. By adopting FFC at all scale levels, we enable the encoder to propagate features at early stages and thus address the issue of generating invalid features inside the holes, helping improve the results as shown in the ablation study in Tab.~\ref{tab:ablation}. 



\begin{figure*}[t]
	\centering
	\includegraphics[width=1.\linewidth]{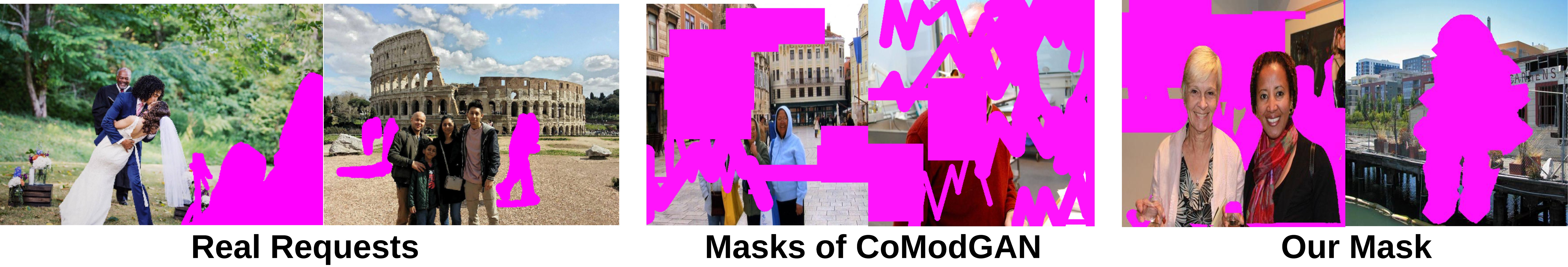}
	\caption{
	\small
	Examples of our object-aware masks generated for training (right) in comparison to the real inpainting requests (left) and the masks generated by CoModGAN~\cite{comodgan} (middle). Note that our masks are more consistent with real user requests.
	}
	\label{fig:mask}
	\vspace*{-.15in}
\end{figure*}

\subsection{Object-aware Training}
\label{sec:mask_scheme}
The algorithm to generate masks for training is crucial. In essence, the sampled masks should be similar to the masks drawn in realistic use-cases.
Moreover, the masks should avoid covering an entire object or most of any new object to discourage model from generating object-like patterns.
Previous works generate mask with square-shaped masks~\cite{pathak2016context,iizuka2017globally} or use random strokes~\cite{partial_conv} or a mixture of both~\cite{deepfillv2,comodgan} for training. The oversimplified mask schemes may cause casual artifacts such as suspicious objects or color blobs.


To better support realistic object removal use cases while preventing the model from trying to synthesize new objects inside the holes, we propose an object-aware training scheme that generate more realistic masks during training as shown in \cref{fig:mask}. 
Specifically, we first pass the training images to PanopticFCN \cite{panopticFCN} to generate highly accurate instance-level segmentation annotations. Next, we sample a mixture of free-form holes~\cite{comodgan} and object holes as the initial mask. Finally, we compute the overlapping ratio between a hole and each instance from an image. If the overlapping ratio is larger than a threshold, we exclude the foreground instance from the hole. Otherwise, the hole is unchanged to mimic object completion. We set the threshold to $0.5$. We dilate and translate the object masks randomly to avoid overfitting. We also dilate the hole on the instance segmentation boundary to avoid leaking background pixels near the hole into the inpainting region.


\subsection{Training Objective and Masked-$R_1$ Regularization}
\label{sec:masked_r1}
Our model is trained with a combination of adversarial loss~\cite{comodgan} and segmentation-based perceptual loss~\cite{lama}. The experiments show that our method can also achieve good results when purely using the adversarial loss, but adding the perceptual loss can further improve the performance. 
In addition, we propose a masked-$R_1$ regularization tailored to stabilize the adversarial training for the inpainting task. 
Different from~\cite{comodgan,lama} that naively apply $R_1$ regularization~\cite{mescheder2018training}, we leverage the mask $m$ to avoid computing the gradient penalty outside the mask, specifically:
\begin{align}
\label{eq:masked_r1}
    \bar{R}_1 = \frac{\gamma}{2} \displaystyle  \E_{p_{\rm{data}}}[\norm{m \odot \nabla D(x)}^2],
\end{align}
where $m$ is the mask indicating the hole region and $\gamma$ is a balancing weight.
The new loss eliminates the potential harmful impact of computing gradient on real pixels, and therefore stabilizes the training.

\begin{table}[]
	\caption{
	\small
        Quantitative evaluation of inpainting on the Places evaluation set. We report FID~\cite{fid}, LPIPS~\cite{lpips}, U-IDS~\cite{comodgan} and P-IDS~\cite{comodgan} scores. 
	}
	\centering
	\resizebox{\columnwidth}{!}{
	\begin{tabular}{l|c|c|c|c|l|c|c|c|c}
	\toprule
		Methods & FID$\downarrow$ & LPIPS$\downarrow$ & U-IDS$\uparrow$ & P-IDS$\uparrow$ & Methods & FID$\downarrow$ & LPIPS$\downarrow$ & U-IDS$\uparrow$ & P-IDS$\uparrow$\\
		\hline
        \B CM-GAN & \B 1.628 & \B{0.189} & \B 37.42 & \B 20.96 & DS~\cite{diverse} & 16.003 & 0.399 & 10.72 & 0.47 \\
        CoModGAN\cite{comodgan} & 3.724 & 0.229 & 32.38 & 14.68 & EC~\cite{edgeconnect} & 12.086 & 0.414 & 9.21 & 0.28 \\
		Lama~\cite{lama} & 3.864 & 0.195 & 29.57 & 10.08 & ICT~\cite{ict} & 16.405 & 0.424 & 8.12 & 0.25 \\
        ProFill~\cite{profill} & 7.700 & 0.230 & 21.19 & 3.87 & HiFill~\cite{hifill} & 37.484 & 0.336 & 9.86 & 0.96\\
        CRFill~\cite{crfill} & 9.657 & 0.233 & 22.90 & 5.53 & SF~\cite{structureflow} & 28.252 & 0.489 & 6.05 & 0.13\\
        DeepFillv2~\cite{deepfillv2} & 13.597 & 0.371 & 14.23 & 1.67 & MEDFE~\cite{MEDFE} & 35.454 & 0.445 & 7.10 & 0.35\\
		\bottomrule
	\end{tabular}
	}
	\label{tab:main}
\end{table}

\section{Experiments}
\noindent \textbf{Implementation Details.} \quad
We conduct the image inpainting experiment at resolution $512\times 512$ on the Places2 dataset~\cite{zhou2017places}.
Our model is trained with Adam optimizer~\cite{adam}. The learning rate and batch size are set to 0.001 and 32, respectively. Our network takes the resized image as input, so that the model can predict the global structure of an image. We apply flip augmentation to increase the training samples.

\begin{figure*}[t]
    \centering
    {\renewcommand{\arraystretch}{0}
    \begin{tabular}{c c}
    \begin{subfigure}[b]{.5\columnwidth}
        \centering
        \includegraphics[width=\columnwidth]{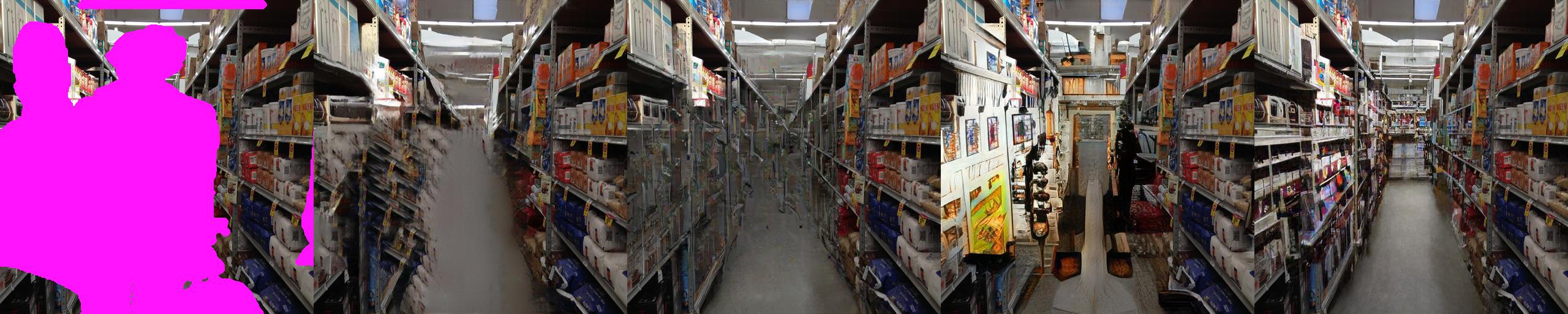}%
    \end{subfigure}&
    \begin{subfigure}[b]{.5\columnwidth}
        \centering
        \includegraphics[width=\columnwidth]{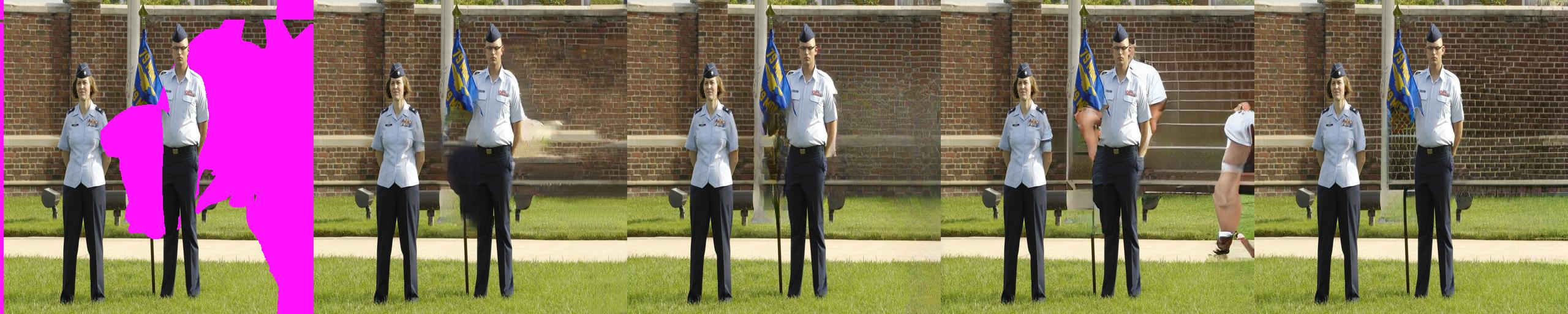}%
    \end{subfigure}\\
    \begin{subfigure}[b]{.5\columnwidth}
        \centering
        \includegraphics[width=\columnwidth]{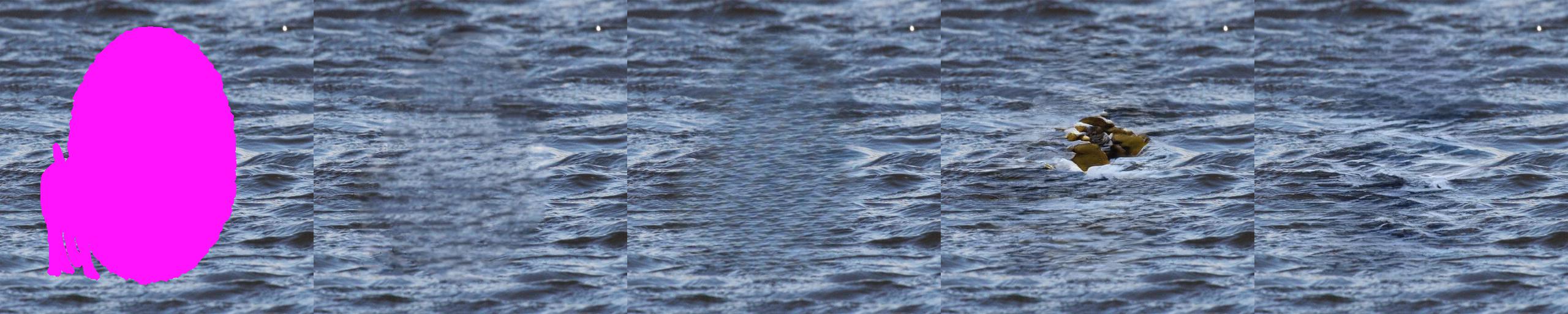}%
    \end{subfigure}&
    \begin{subfigure}[b]{.5\columnwidth}
        \centering
        \includegraphics[width=\columnwidth]{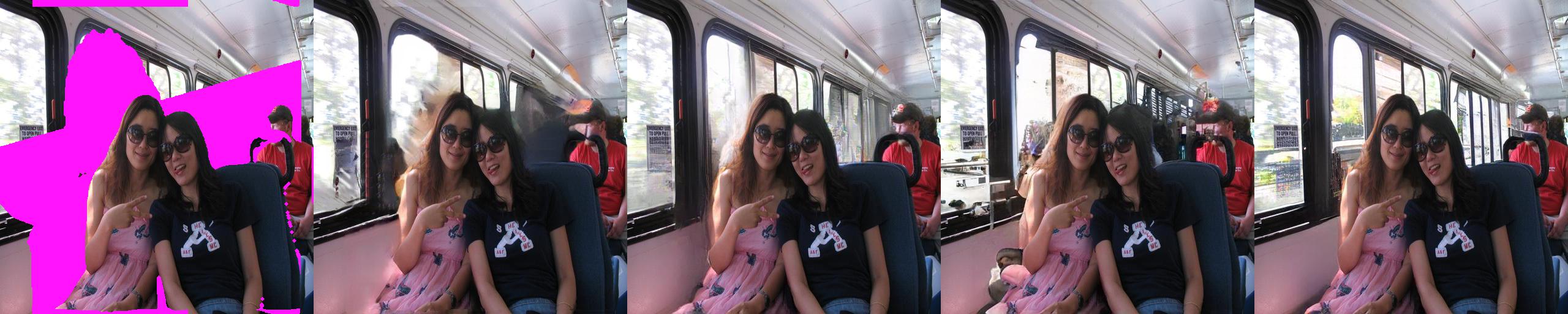}%
    \end{subfigure}\\
    \begin{subfigure}[b]{.5\columnwidth}
        \centering
        \includegraphics[width=\columnwidth]{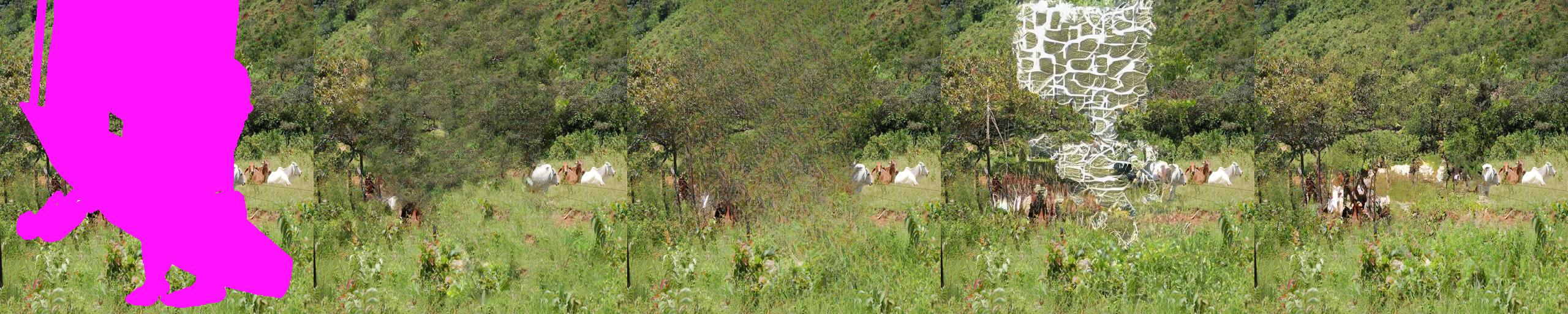}%
    \end{subfigure}&
    \begin{subfigure}[b]{.5\columnwidth}
        \centering
        \includegraphics[width=\columnwidth]{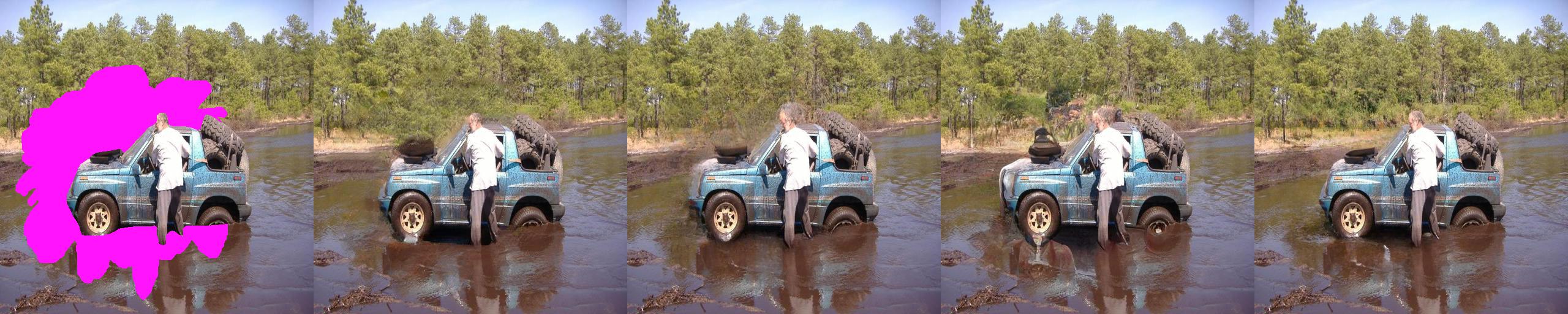}%
    \end{subfigure}\\
    \begin{subfigure}[b]{.5\columnwidth}
        \centering
        \includegraphics[width=\columnwidth]{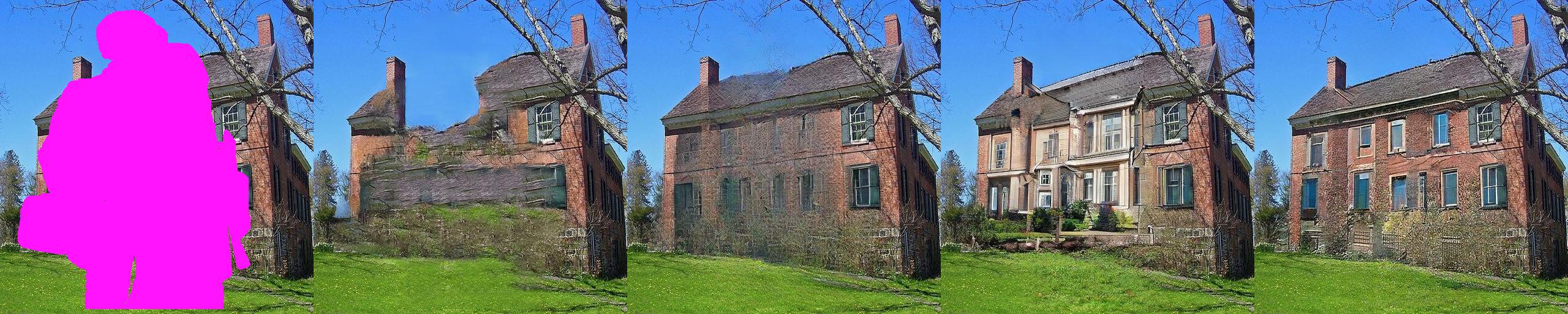}%
    \end{subfigure}&
    \begin{subfigure}[b]{.5\columnwidth}
        \centering
        \includegraphics[width=\columnwidth]{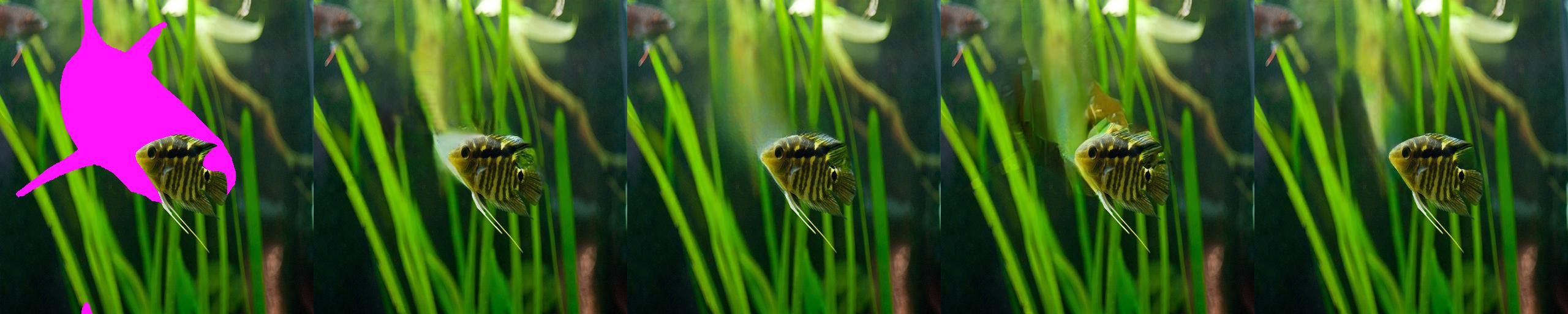}%
    \end{subfigure}\\
    \begin{subfigure}[b]{.5\columnwidth}
        \centering
        \includegraphics[width=\columnwidth]{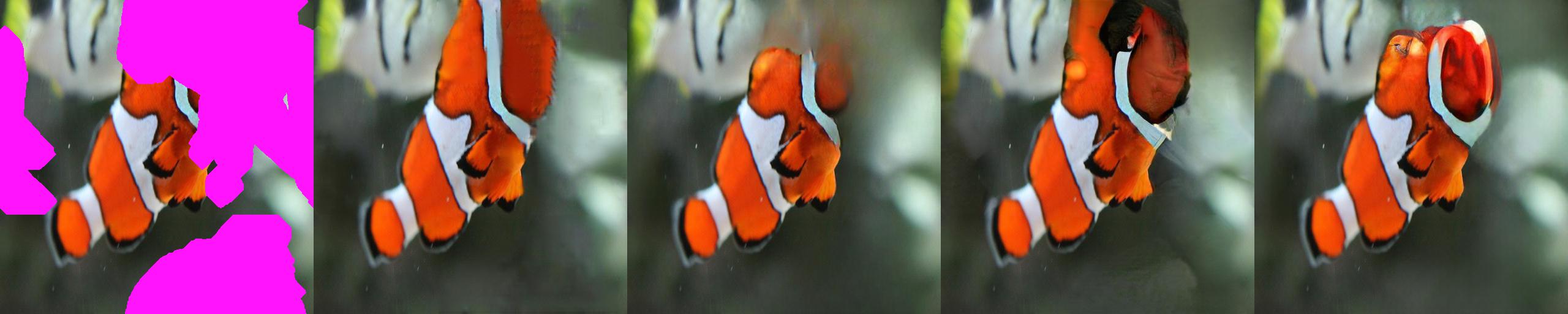}%
    \end{subfigure}&
    \begin{subfigure}[b]{.5\columnwidth}
        \centering
        \includegraphics[width=\columnwidth]{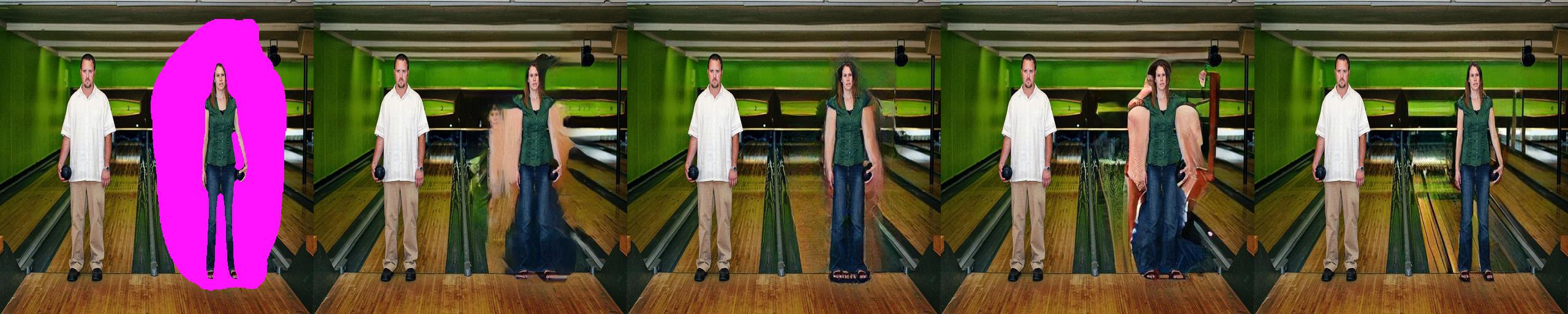}%
    \end{subfigure}\\
    \begin{subfigure}[b]{.5\columnwidth}
        \centering
        \includegraphics[width=\columnwidth]{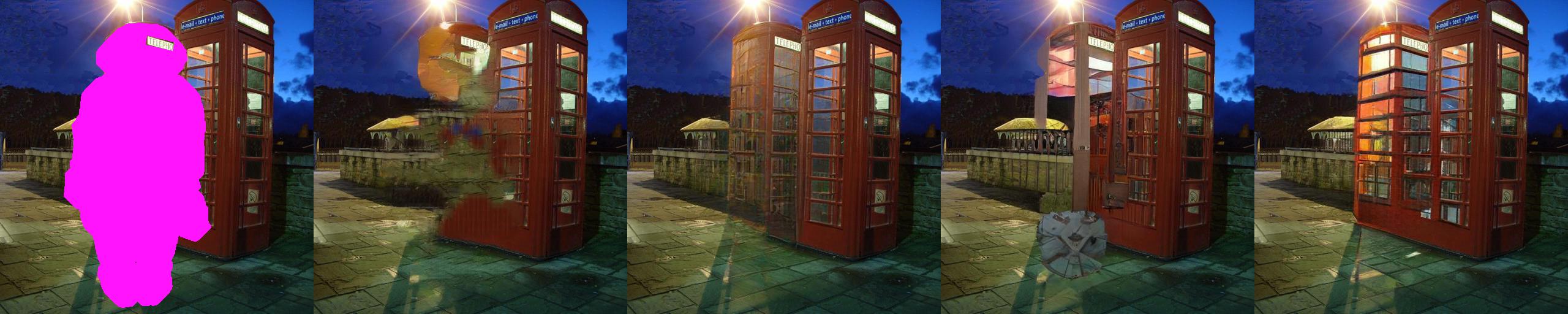}%
    \end{subfigure}&
    \begin{subfigure}[b]{.5\columnwidth}
        \centering
        \includegraphics[width=\columnwidth]{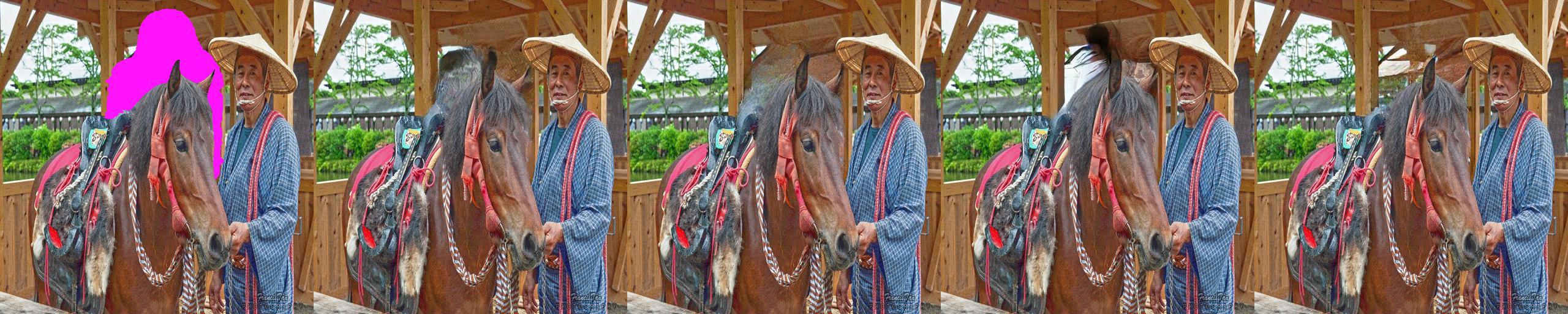}%
    \end{subfigure}\\
    \begin{subfigure}[t]{.5\columnwidth}
        \centering 
        \includegraphics[width=\textwidth]{figure_compressed/caption.png}%
    \end{subfigure}&
    \begin{subfigure}[t]{.5\columnwidth}
        \centering 
        \includegraphics[width=\columnwidth]{figure_compressed/caption.png}%
    \end{subfigure}
    \vspace{-0.1cm}
    \end{tabular}}
    \caption[]
    {\small Visual comparisons on Places2 with our synthesized masks including large random masks (left) and masks for the distractor removal scenario (right). We show
    the input images and the results of ProFill~\cite{profill},
    Lama~\cite{lama}, CoModGAN~\cite{comodgan} and CM-GAN (ours). Best viewed by zoom-in on screen.} 
    \label{fig:main_compare}
\end{figure*}

\begin{figure*}[t]
    \centering
    {\renewcommand{\arraystretch}{0}
    \begin{tabular}{c c}
    \begin{subfigure}[b]{.5\columnwidth}
        \centering
        \includegraphics[width=\columnwidth]{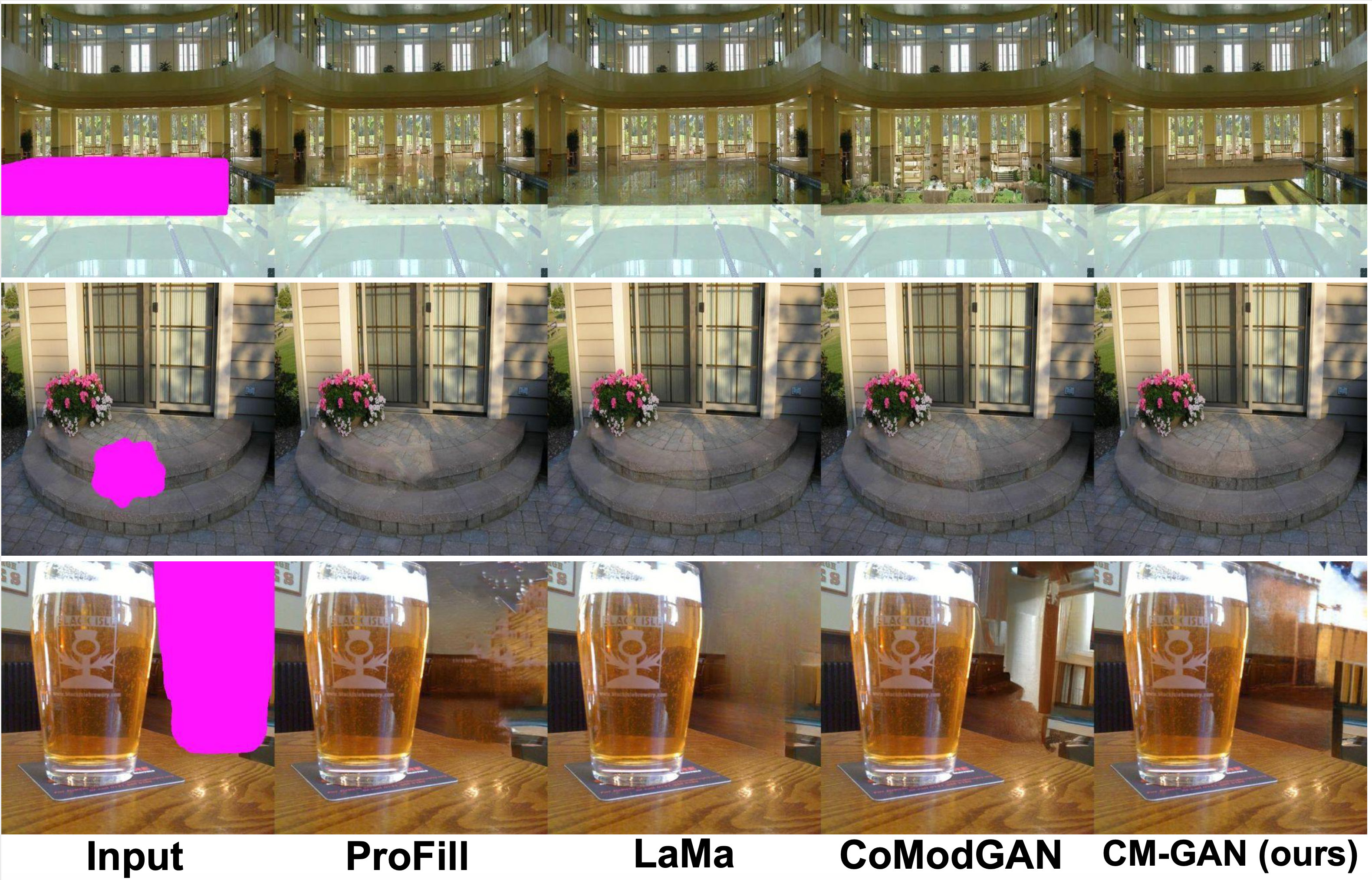}%
    \end{subfigure}&
    \begin{subfigure}[b]{.5\columnwidth}
        \centering
        \includegraphics[width=\columnwidth]{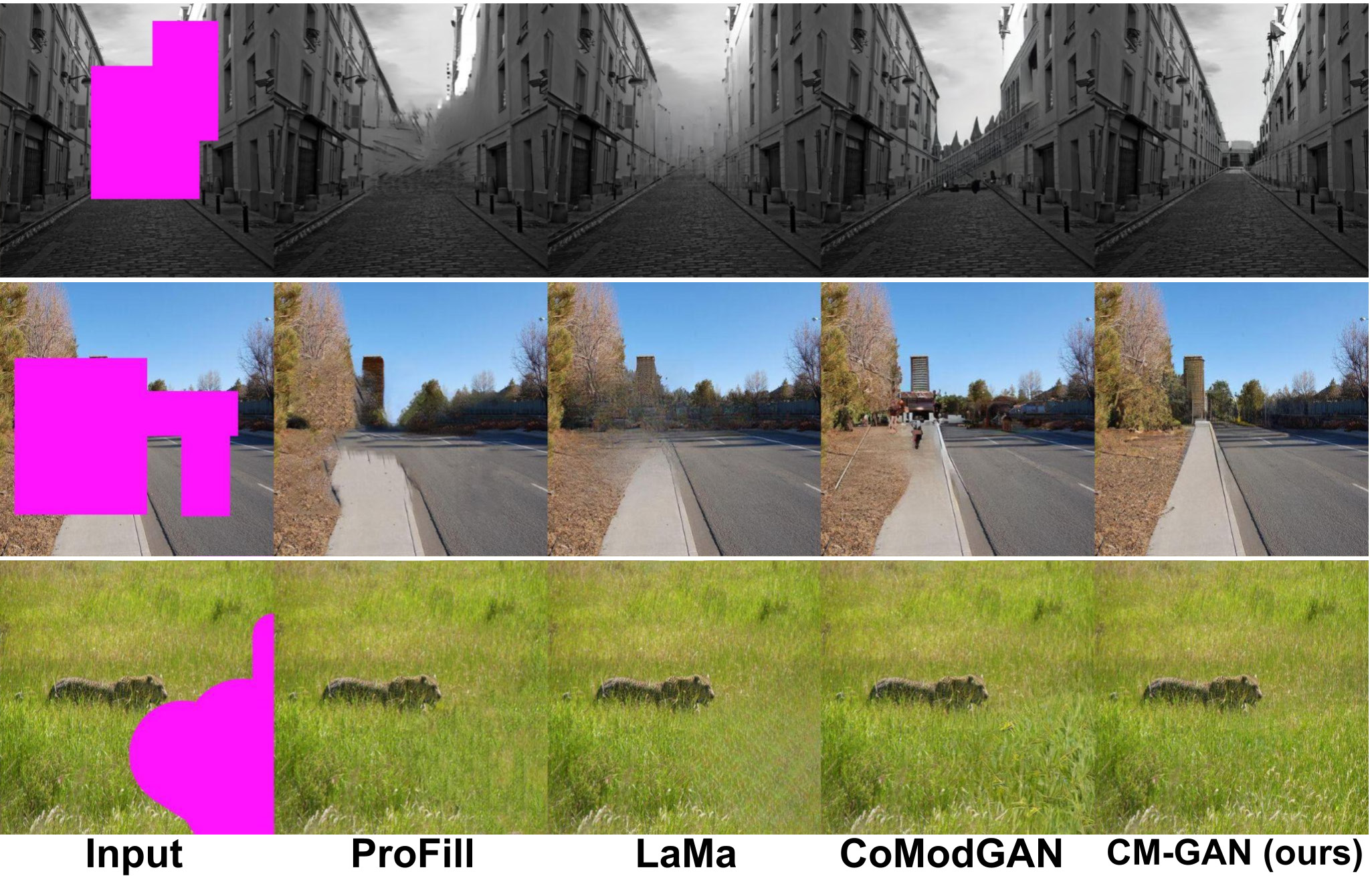}%
    \end{subfigure}\\
    \end{tabular}}
    \caption[]
    {\small 
    Visual comparisons on Places2 with the masks proposed by Lama~\cite{lama}.
    } 
    \label{fig:lama_compare}
\end{figure*}




\noindent \textbf{Evaluation Metrics.} \quad
We report the numerical metrics on the validation set of Places2 which contains 36.5k images. For the numerical evaluation, we compute \emph{Frchet Inception Distance} (FID)~\cite{fid} and \emph{Perceptual Image Patch Similarity Distance} (LPIPS)~\cite{lpips}. We also adopt the \emph{Paired/Unpaired Inception Discriminative Score} (P-IDS/U-IDS)~\cite{comodgan} for evaluation.

\subsection{Comparisons to Existing Methods}
We set channel numbers of our network to have a similar model capacity as CoModGAN and LaMa as shown in \cref{tab:model_size}.\\
\noindent \textbf{Quantitative Evaluation.} \quad
\cref{tab:main} presents the comparison of our method against a number of recent methods using our masks.
Results show that our method significantly outperforms all other methods in terms of FID, LPIPS, U-IDS and P-IDS.
We notice that with the assist of perceptual loss, LaMa~\cite{lama} and our CM-GAN achieve significantly better LPIPS score than CoModGAN and other methods, attributing to additional semantic guidance provided of the pre-trained perceptual model.
Compared to LaMa, our CM-GAN reduces FID by over 50\% from 3.864 to 1.628, which can be explained by the typically blurry results of LaMa versus ours which tend to be sharper.

We evaluate generalization of CM-GAN to other types of masks including the wide mask~\cite{lama} and the mask of CoModGAN~\cite{comodgan}. 
We also fine-tune CM-GAN with masks of \cite{lama} and \cite{comodgan} (denoted by CM-GAN\ding{61}) and report the results.
As shown in \cref{tab:lama_mask}, our models with and without fine-tuning achieve clear performance gain and demonstrate its  generalization ability. Notably, CM-GAN trained on our object-aware masks outperforms CoModGAN on the CoModGAN mask, confirming the better generation capacity of CM-GAN.
The strong capacity of CM-GAN brings further performance gain after fine-tuning.

\noindent \textbf{Qualitative Evaluation.} \quad
\cref{fig:main_compare}, \cref{fig:lama_compare} and \cref{supp:fig:comodgan-mask} presents visual comparisons of our method with state-of-the-art methods on our synthesized masks and other types of mask introduced by~\cite{lama} and~\cite{comodgan}, respectively. ProFill~\cite{profill} generates incoherent global structures such as the smoothed building and tends to bleed color to the background for the object-removal case. CoModGAN~\cite{comodgan} produces structural artifacts and color blobs. LaMa~\cite{lama} is superior on repeating structures, but tends to generate blurry results on large holes, especially on nature scenes.
In contrast, our method produces more coherent semantic structures and hallucinates cleaner textures for various types of scenarios. 

\subsection{Ablation Study}
We perform a set of ablation experiments to show the importance of each component of our model. All ablated models are trained and evaluated on the Places2 dataset. Results of the ablations are shown in \cref{tab:ablation}. Below we describe the ablations from the following aspects:

\noindent \textbf{Masked-$R_1$ regularization} \quad
We start from a \texttt{baseline} with a simple encoder-decoder structure based on global-vector modulation~\cite{comodgan} and skip connection. We compare the baseline trained with $R_1$ regularization with the model trained with masked $R_1$ regularization regularization (\texttt{baseline+m$R_1$}). From the result, the masked $R_1$ regularization improves the numerical metrics as the designed loss avoids computing gradient at the fixed input region.

\begin{figure}[t]
	\centering
	\includegraphics[width=1.0\linewidth]{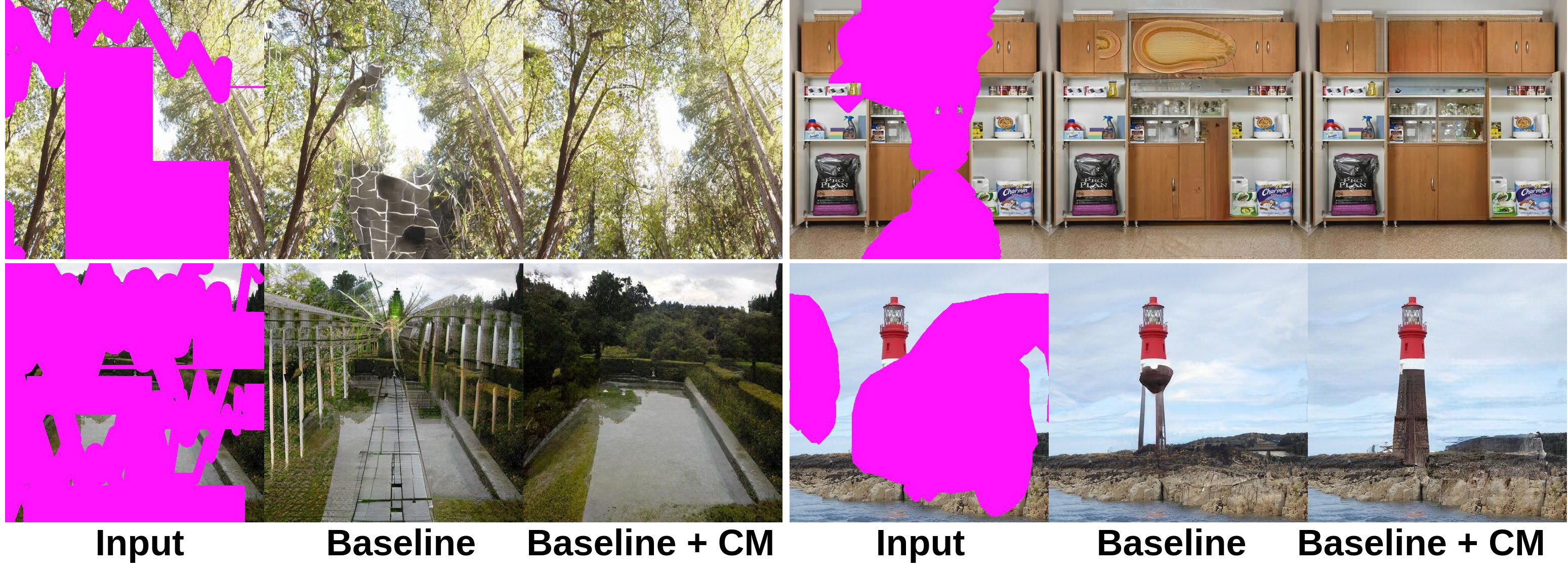}
	\caption{
    \small
	Compared to a global modulation baseline, 
	CM significantly improves the coherence of the synthesized color, textures, global structures and objects.
	}
	\label{fig:ablation_CM}
\end{figure}

\noindent \textbf{Cascaded Modulation} \quad
We next evaluate the cascaded modulation design on top of the baseline network and m$R_1$ loss. 
Specifically, we evaluate a baseline model with spatial modulation instead of the global code modulation, i.e. \texttt{baseline + s}. The performance improvement verify the effectiveness of the spatial adaptation introduced by spatial modulation.
Next, we cascade global and spatial modulation on the baseline to get the main CM model \texttt{baseline+CM (g-s) ours} which improves all numerical metrics.
To better understand CM, we visually compare \texttt{baseline} with \texttt{baseline+CM (g-s) ours} in \cref{fig:ablation_CM} and find that CM significantly improves the synthesized color, texture and global semantic and corrects the color blob artifact~\cite{comodgan}, which confirms the effectiveness of CM on correcting the incoherent feature in a global-semantic aware fashion.

\noindent \textbf{Choices of Second-stage Modulation} \quad
We evaluate other variant of the second-stage modulation choices: i) we replace our StyleGAN2-compatible spatial with skip connection, resulting in a model that cascades global modulation twice, i.e. \texttt{baseline+CM(g-g)}, ii) we test the CM baseline with the spatial modulation of \cite{stylemapgan}, i.e. \texttt{baseline+CM(g-\cite{stylemapgan})} and iii) we drop the demodulation step (\cref{eq:demodulation}) from the spatial modulation step, resulting a model with a plain spatial modulation operation, i.e. \texttt{Baseline + CM (g-s) plain}.
From the results, our spatial modulation outperforms the global modulation version as we modulate feature using both global and spatial code. We found \cite{stylemapgan} does not improve CM as the instance normalization of \cite{stylemapgan} is designed for StyleGAN and is less compatible with our baseline. Furthermore, demodulation seems crucial to the model as it regularizes the intermediate feature activation. Finally, for the same reason as \cite{stylemapgan}, we found SPADE~\cite{spade} not compatible with our baseline due to the use of batch normalization.

\noindent \textbf{Perceptual Loss} \quad The perceptual loss (perc.) provides additional semantic supervision to the network and can significantly improve the FID metrics. However, it slightly decreases the discriminative U/P-IDS scores as perceptual loss may lead to certain visual patterns that are imperceptible to human.

\noindent \textbf{Fast Fourier Convolutions Encoder} \quad
The Fast Fourier Convolution (FFC) encoder further brings significant performance gains on top of the cascaded modulation and perceptual loss as shown in the table, which validates the importance of more effective encoder with wider receptive fields in early encoding stages.

\noindent \textbf{Object-aware Training} \quad
To study the effect of object-aware training (OT), we retrain LaMa~\cite{lama} and CoModGAN~\cite{comodgan} on our object-aware masks. The results show that object-aware training improves the performance of both of these models consistently. However, our full model still outperforms these retrained models significantly. Notably, our model reduces FID of the retrained CoModGAN (with OT) by 40\% from 2.599 of to 1.628.

\begin{table}[]
	\caption{
	\small
	   Ablation study of  our model design (architecture, loss, training scheme) including masked-$R_1$ loss (m$R_1$), cascaded modulation option (CM), Fourier convolution (FFC), perceptual loss (perc.) and object-aware training (OT).
	   We report FID~\cite{fid}, LPIPS~\cite{lpips}, U-IDS~\cite{comodgan} and P-IDS~\cite{comodgan} scores.
	}
	\centering
    \resizebox{\columnwidth}{!}{
	\begin{tabular}{c|l|c|c|c|c}
	\toprule
		Ablations &\multicolumn{1}{c|}{Methods} & FID$\downarrow$ & LPIPS$\downarrow$ & U-IDS$\uparrow$ & P-IDS$\uparrow$\\
		\hline
		\multirow{2}{*}{\em{\centering Masked-$R_1$(m$R_1$)}}&Baseline & 2.530 & 0.221 & \B36.59 & 21.10 \\ 
		&\B Baseline + m$\boldsymbol{R_1}$ & \B2.475 & \B0.221 & 36.58 & \B21.55 \\ 
		\hline
        \multirow{5}{*}{\em {Cascaded Modulation (CM)}} & Baseline + s & 2.398 & 0.218 & 36.72 & 21.94 \\
		& Baseline + CM (g-g) & 2.247 & 0.226 & 37.12 & 22.70 \\ 
		& Baseline + CM (g-\cite{stylemapgan}) & 2.915 & \B0.221 & 35.72 & 20.44 \\ 
		& Baseline + CM (g-s) plain & 2.392 & 0.224 & 37.23 & 22.67 \\
		& \B Baseline + CM (g-s) & \B 2.187 & 0.225 & \B37.76 & \B23.86  \\ 
		\hline
        \multirow{2}{*}{\em {FFC and perc.}} & Baseline + CM + perc. & 1.730 & 0.195 & 36.12 & 19.73\\ 
        & \B Baseline + CM + perc. + FFC & \B 1.628 & \B 0.189 & \B 37.42 & \B 20.96\\
        \hline
        \multirow{5}{*}{\em{Object-aware Training (OT)}}
        & CoModGAN~\cite{comodgan} & 3.724 & 0.229 & 32.38 & 14.68\\ & CoModGAN~\cite{comodgan} + OT & 2.599 & 0.222 & 35.45 & 20.08 \\ 
		& Lama~\cite{lama} & 3.864 & 0.195 & 29.57 & 10.08 \\
		& Lama~\cite{lama} + OT & 2.884 & 0.192 & 31.32 & 15.10 \\
		& \B CM-GAN (full) & \B 1.628 & \B 0.189 & \B 37.42 & \B 20.96\\
		\bottomrule
	\end{tabular}
	}
	\label{tab:ablation}
\end{table}

\begin{table}[]
	\caption{
	    \small
	    Generalization evaluation on other types of mask including wide masks~\cite{lama} and CoMoGAN masks~\cite{comodgan}.
	    CMGAN\ding{61} are models fine-tuned on the two types of mask.
	}
	\centering
	\resizebox{\columnwidth}{!}{
	\begin{tabular}{l|c|c|c|c|l|c|c|c|c}
	\toprule
	    \multicolumn{5}{c|}{\B \em Wide masks~\cite{lama}} & \multicolumn{5}{c}{\B \em CoModGAN masks ~\cite{comodgan}}\\
	    \hline
		Methods & FID$\downarrow$ & LPIPS$\downarrow$ & U-IDS $\uparrow$ & P-IDS $\uparrow$ & Methods & FID$\downarrow$ & LPIPS$\downarrow$ & U-IDS $\uparrow$ & P-IDS $\uparrow$\\
		\hline
        \underline{CM-GAN} & \underline{1.521} & 0.129  & \underline{39.24} & \underline{23.24} & \underline{CM-GAN} & \underline{6.811} & \underline{0.313} & \underline{26.13} & \underline{10.84} \\
        \B CM-GAN\ding{61} & \B 1.329 & \underline{0.126} & \B 40.20 & \B 25.59 & \B CM-GAN\ding{61} & \B 5.863 & \B 0.310 & \B 27.92 & \B 12.45 \\
        \hline
		LaMa~\cite{lama}& 1.838 & \B0.123 & 35.00 & 15.12 & CoModGAN\cite{comodgan} & {7.790} & {0.344} & {24.87} & {10.47}\\
        CoModGAN\cite{comodgan}& 1.964 & 0.140 & 37.69 & 21.42 & LaMa~\cite{lama} & 12.442 & {0.316} & 18.71 & 4.36 \\
        ProFill~\cite{profill}& 3.333 & 0.142 &  29.12 & 8.39 & ProFill~\cite{profill}	& 20.314 & 0.352 & 11.21 & 1.23 \\
		\bottomrule
	\end{tabular}
	}
	\label{tab:lama_mask}
\end{table}

\begin{table}[]
	\caption{
	\small
	    The inference complexities. Our model has a similar number of parameters and FLOPs as other recent models.
	}
	\centering
	\begin{tabular}{l|c|c|c}
	\toprule
		Models & \#Params of $\mathcal{G}$ & \#Params of $\mathcal{D}$ & FLOPs\\
		\hline
		CoModGAN~\cite{comodgan}& 79.79M & 28.98M & 345.54G\\
		LaMa~\cite{lama}& 51.25M & 9.258M & 395.30G\\
		CM-GAN (ours) & 75.28M & 28.98M & 373.60G\\
		\bottomrule
	\end{tabular}
	\label{tab:model_size}
\end{table}

\vspace{-5mm}
\subsection{User Study}
We conduct a user study to better evaluate the visual quality of our method. Specifically, we generate samples for evaluation using the Places2 evalation set and three types of masks: our object-aware mask, wide mask~\cite{lama} and mask from real user request. The former data class contains 30 samples while the latter contains 13 real inpainting requests online following~\cite{crfill}. Each input image with the region to be removed and the results of different methods are presented to online users who are asked to select the best result. Finally, we collect votes from all users. Results in \cref{tab:user_study} shows that our method receives the majority of votes on both the synthetic data and realistic object removal requests.
\begin{table}[]
\caption{
\small
The user study. For each mask type, we show the number of votes and percentages for different methods (ProFill, LaMa, CoModGAN, and ours)
}
\centering
\begin{tabular}{l|c|c|c|c}
\toprule
Masks&ProFill~\cite{profill}&Lama~\cite{lama}&CoModGAN~\cite{comodgan}&CM-GAN\\\hline
Our mask&20 (5\%) & 68 (17\%) & 83 (20\%) & \textbf{234 (58\%)}\\
Wide mask&45 (10\%) &107 (25\%) &94 (22\%) & \textbf{186 (43\%)}\\
User mask&15 (5\%) &101 (35\%) &55 (19\%) & \textbf{120 (41\%)}\\
\bottomrule
\end{tabular}
\label{tab:user_study}
\end{table}

\begin{figure*}[]
	\centering
    \includegraphics[width=1\linewidth]{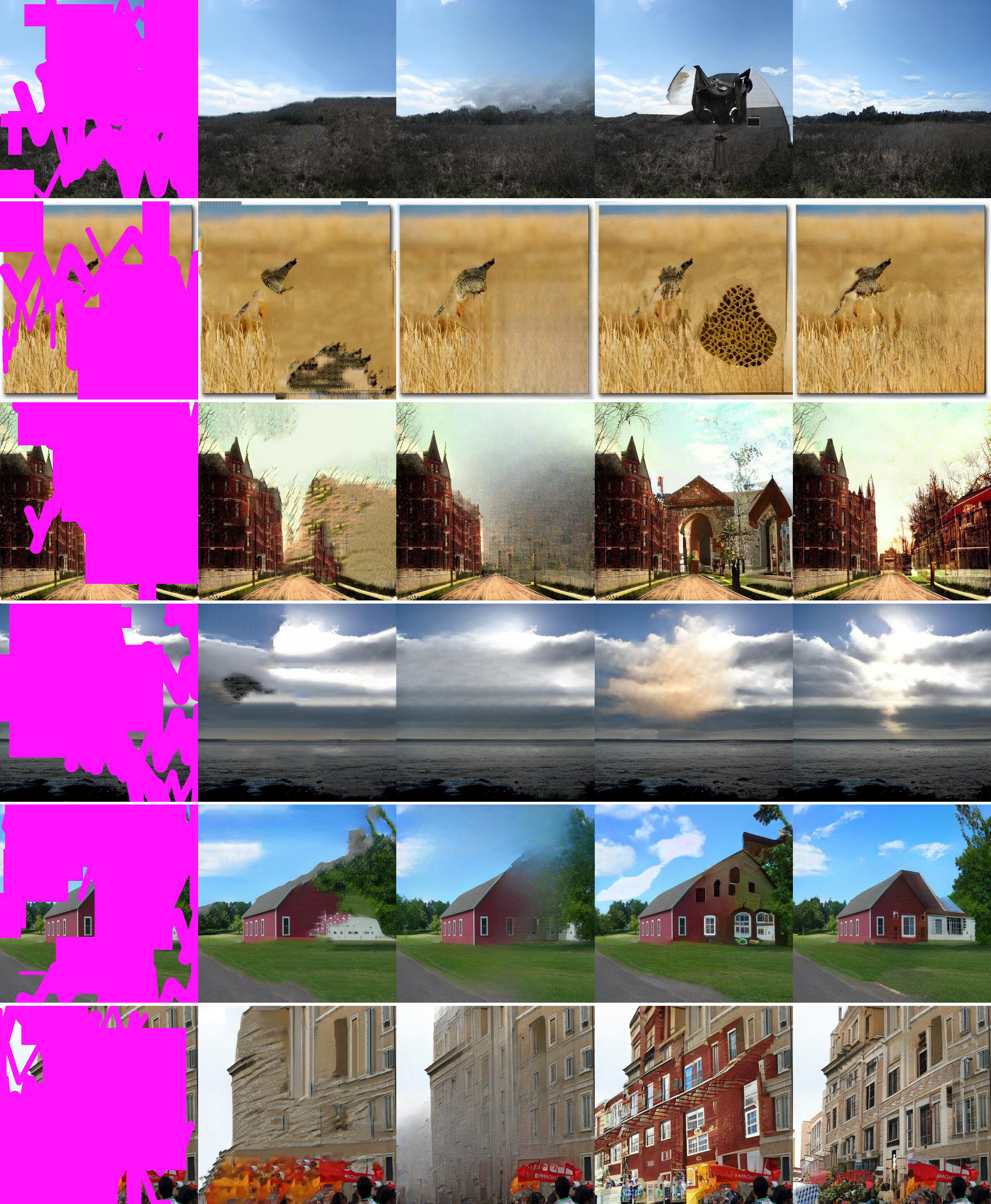}
	\includegraphics[width=1\linewidth]{figure_compressed/caption.png}
	\caption{
	\small
	{
	\textbf{Visual comparison on the mask of CoModGAN~\cite{comodgan}. 
	Best viewed by zoom-in on screen.}
	}
	}
	\label{supp:fig:comodgan-mask}
\end{figure*}

	\vspace{-5mm}
\section{Conclusion}
In this paper, we present a new approach tailored to real-world image inpainting. Our method is based on a new modulation block that cascades global modulation with spatial modulation for better pass global context into the hole region. We further propose a training scheme based on object-aware mask sampling to improve generalization to real use cases. Finally, we propose specifically designed masked $R_1$ regularization to stabilize the adversarial training of image inpainting networks.
Our method achieves the new state-of-the-art performance on the Places2 dataset and better visual quality. 

Currently, our model is still limited in synthesizing large objects like humans or animals. One possible solution is to train a specialist inpainting model for specific types of objects. Another direction is to leverage depth and semantic segmentation for more precise structure-aware inpainting.



	\section*{Acknowledgement}
	We would like to thank Qing Liu for the efforts and help on the demo interface and other experiments.
	
	\clearpage
	\appendix
	 \setcounter{section}{0}
{
 \centering
 \large \textbf{Appendix} \\
}
\vspace{3mm}

We provided more analysis, visual results and implementation details
in the appendix,
including
analysis of object-aware training (\cref{supp:sec:object-training}), effect of the masked-$R_1$ regularization (\cref{supp:sec:masked-r1}), visual comparison to other methods (\cref{supp:sec:visual-results}), visual comparison on other types of masks (\cref{supp:sec:other-mask}) and more implementation details (\cref{supp:sec:implementation-details}).
We also introduce comparisons to the recent transformer-based approach TFill~\cite{zheng2021tfill}.
All visual results are in high-resolution and best viewed by zoom-in on screen.

\vspace{-8mm}
\section{Visual Effects of Object-aware Training on Other Models}
\label{supp:sec:object-training}
\begin{figure*}[]
	\centering
	\includegraphics[width=1\linewidth]{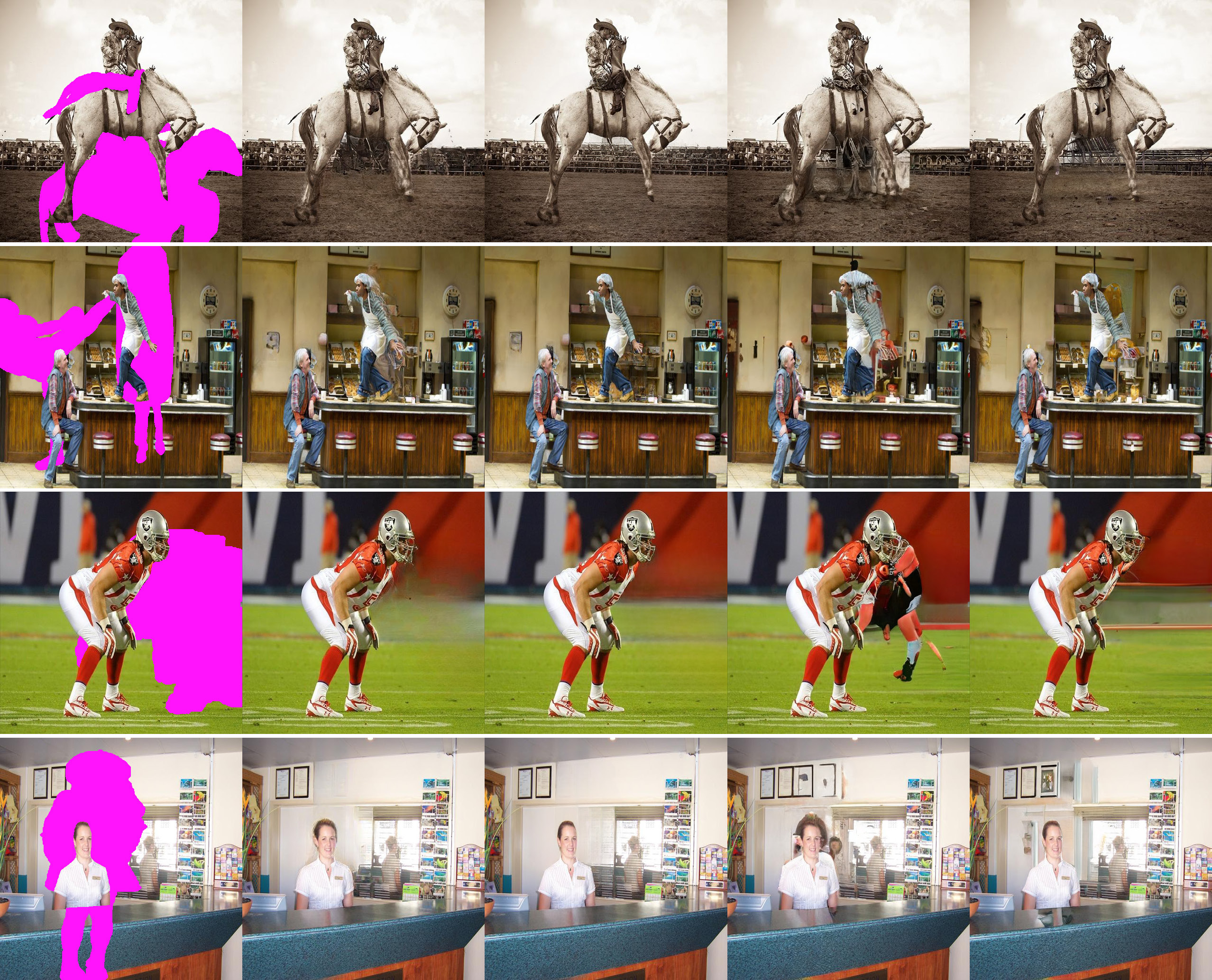}
	
	\includegraphics[width=1\linewidth]{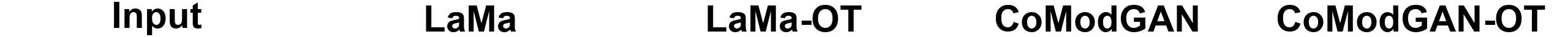}
	
	\caption{
		\small{
			\textbf{
				The effect of object-aware training on other models.} Object-aware training (OT) improves other models including LaMa~\cite{lama} and CoModGAN~\cite{comodgan} on achieving sharper boundaries and clearer background. Best viewed by zoom-in on screen.
		}
	}
	\label{supp:fig:OT-effect}
\end{figure*}

\begin{figure*}[]
	\centering
	
	\includegraphics[width=1\linewidth]{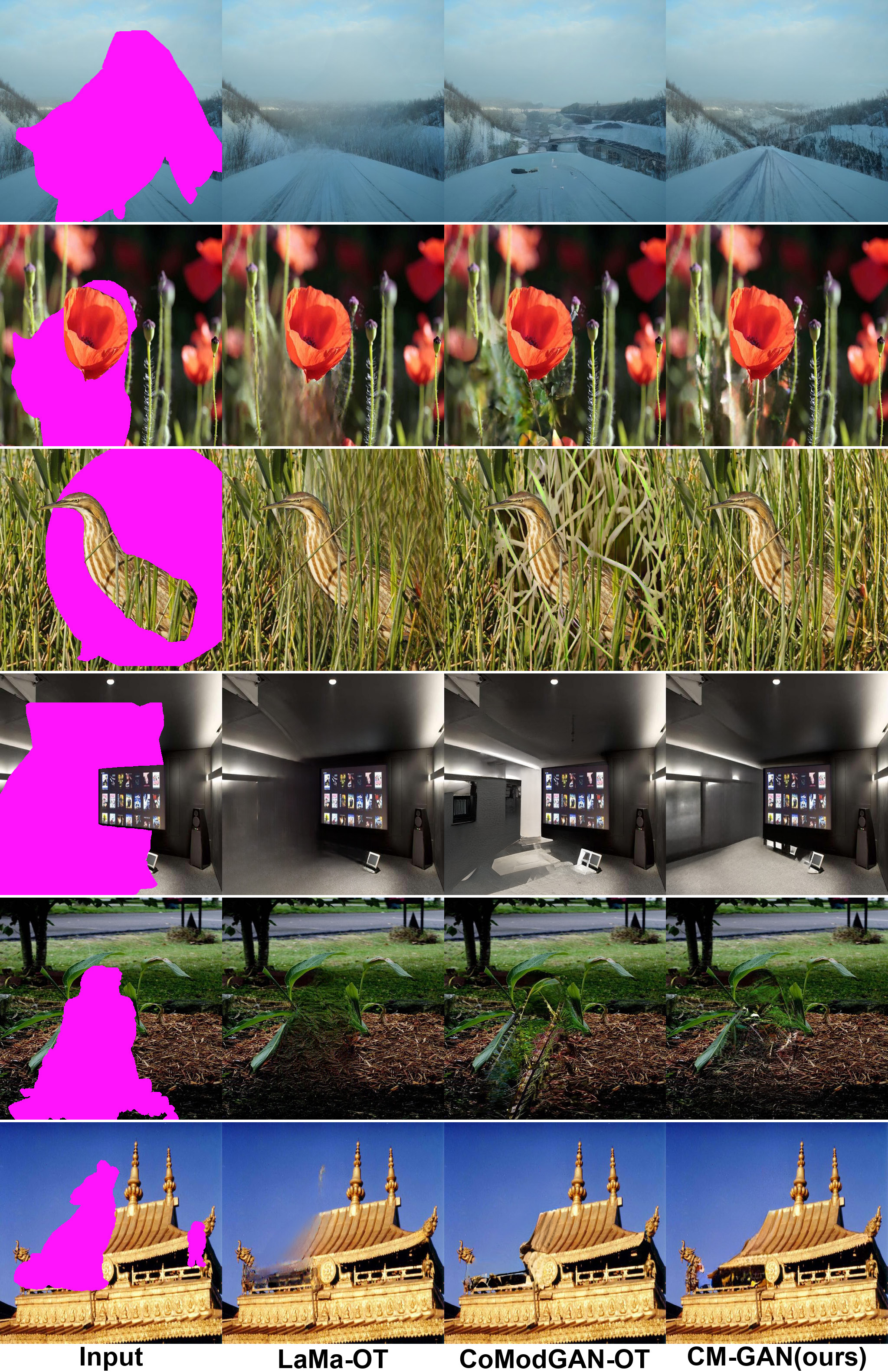}
	
	\caption{
		\small
		{
			\textbf{Results of CM-GAN in comparison to LaMa~\cite{lama} and CoModGAN~\cite{comodgan} trained with object-aware training (OT).}
			Best viewed by zoom-in on screen.
		}
	}
	\label{supp:fig:OT-compare}
\end{figure*}

To analyze the generalization of object-aware training to other recent inpainting methods~\cite{lama,comodgan} while complementing the numerical results in Table 2 of the main paper, we provide the supplemental visual effect of object-aware training in \Cref{supp:fig:OT-effect} and \Cref{supp:fig:OT-compare}.

\Cref{supp:fig:OT-effect} presents the visual comparison of LaMa~\cite{lama} and CoModGAN~\cite{comodgan} trained without or with object-aware training (OT). Object-aware training in general improves other state-of-the-art models including LaMa and CoModGAN on retaining object boundaries and background under the distractor removal scenario.

Furthermore, \Cref{supp:fig:OT-compare} presents the visual comparison of our method and the state-of-the-art models trained with object-aware training, including LaMa-OT and CoModGAN-OT.
CM-GAN with object-aware training achieves better performance than other state-of-the-art models trained with object-aware training, validating the strong generation capacity of CM-GAN.

\section{The Effect of The Masked-$R_1$ Regularization}
\label{supp:sec:masked-r1}
\begin{figure}[]
	\centering
	\includegraphics[width=0.8\linewidth]{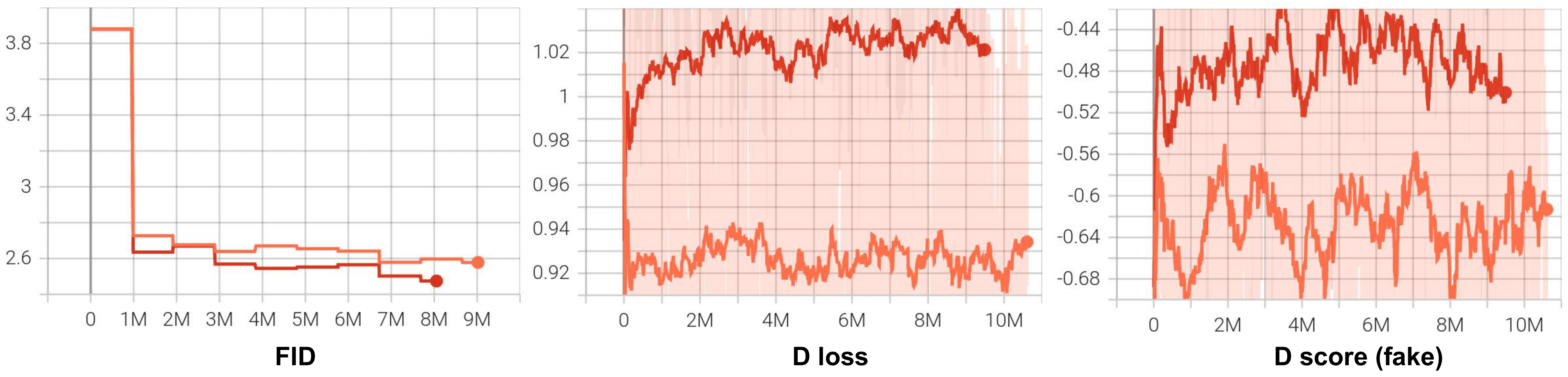}
	\caption{
		\small{
			\textbf{The convergence curves of the baseline models trained with masked-$R_1$ regularization (red) and $R_1$ regularization (orange).} The masked-$R_1$ regularization help the model achieve lower FID scores, higher discriminator classification loss and makes the discriminator harder to distinguish fake samples.
		}
	}
	\label{fig:loss}
\end{figure}

\Cref{fig:loss} visualizes the effect of masked-$R_1$ during the training. Specifically, we visualize the baseline model trained with masked-$R_1$ regularization (red) and $R_1$ regularization (orange). The masked-$R_1$ regularization helps the model achieve lower FID scores, higher discriminator classification loss and makes discriminator harder to distinguish fake samples.


\section{Additional Qualitative Results}
\label{supp:sec:visual-results}
In the following, we provide the supplementary qualitative results of methods evaluated in Table 1 of the main paper. In addition, we include evaluation of a recent transformer-based approach, i.e. TFill~\cite{zheng2021tfill} in \Cref{supp:subsec:transformer}.

\subsection{More Visual Comparisons with ProFill, LaMa, and CoModGAN}
\label{supp:subsec:visual-results}
\begin{figure*}[]
	\centering
	\includegraphics[width=1\linewidth]{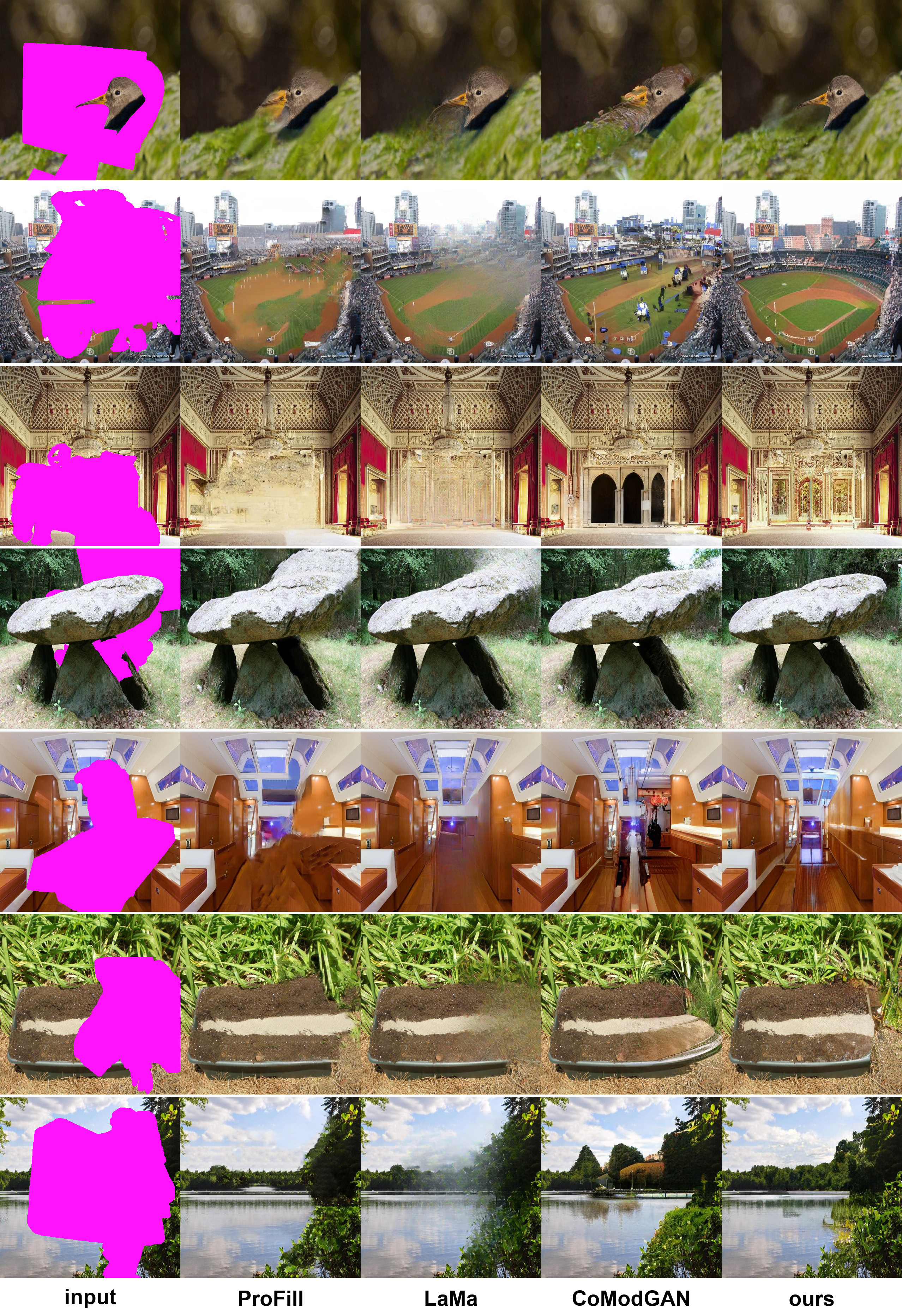}
	
	\caption{
		\small
		{
			\textbf{Results of CM-GAN in comparison to ProFill~\cite{profill}, LaMa~\cite{lama} and CoModGAN~\cite{comodgan}.}
			Best viewed by zoom-in on screen.
		}
	}
	\label{supp:fig:main-compare1}
\end{figure*}

\begin{figure*}[]
	\centering
	\includegraphics[width=1\linewidth]{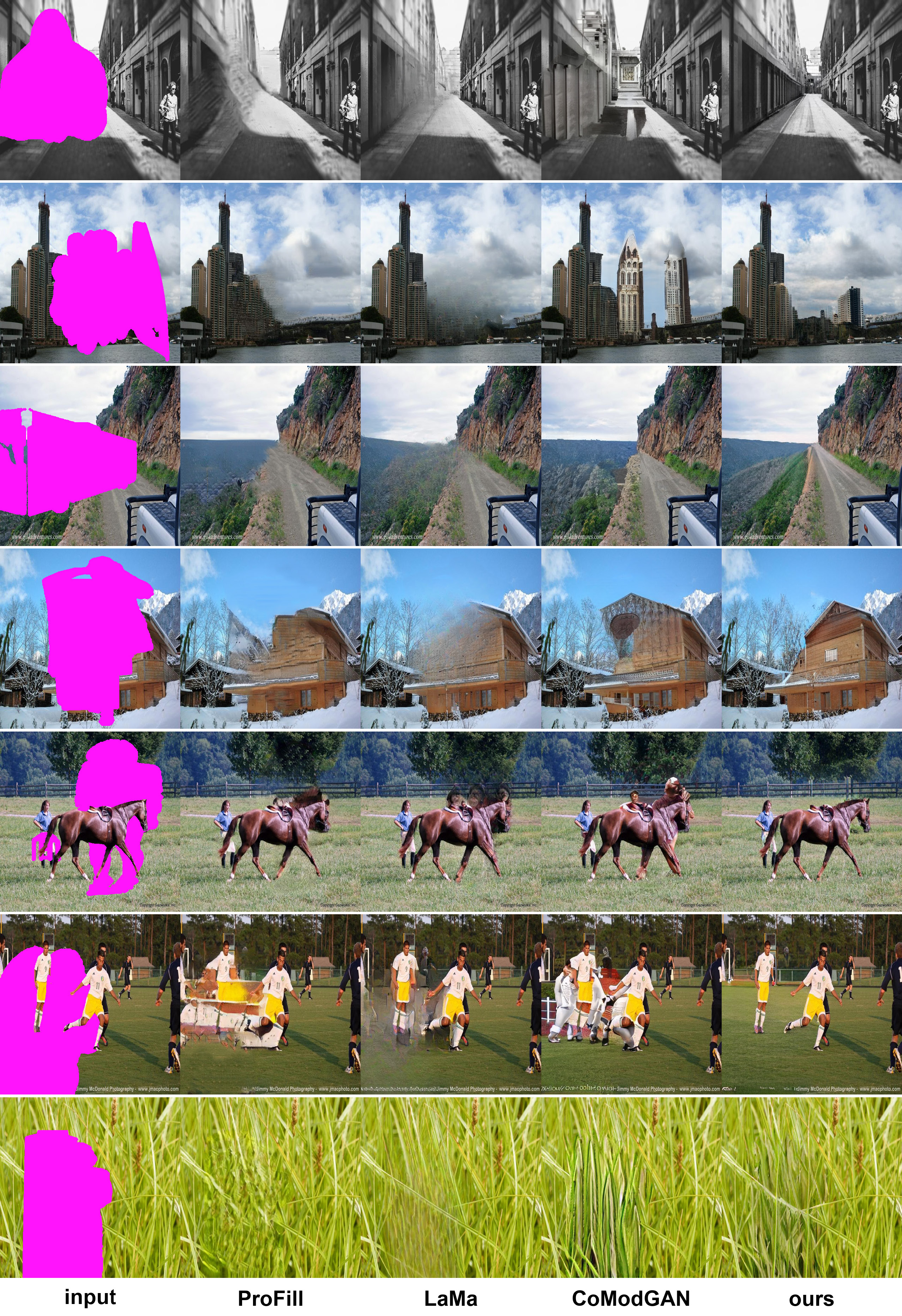}

	\caption{
		\small
		{
			\textbf{Results of CM-GAN in comparison to ProFill~\cite{profill}, LaMa~\cite{lama} and CoModGAN~\cite{comodgan}.}
			Best viewed by zoom-in on screen.
		}
	}
	\label{supp:fig:main-compare2}
\end{figure*}

\begin{figure*}[]
	\centering
	\includegraphics[width=1\linewidth]{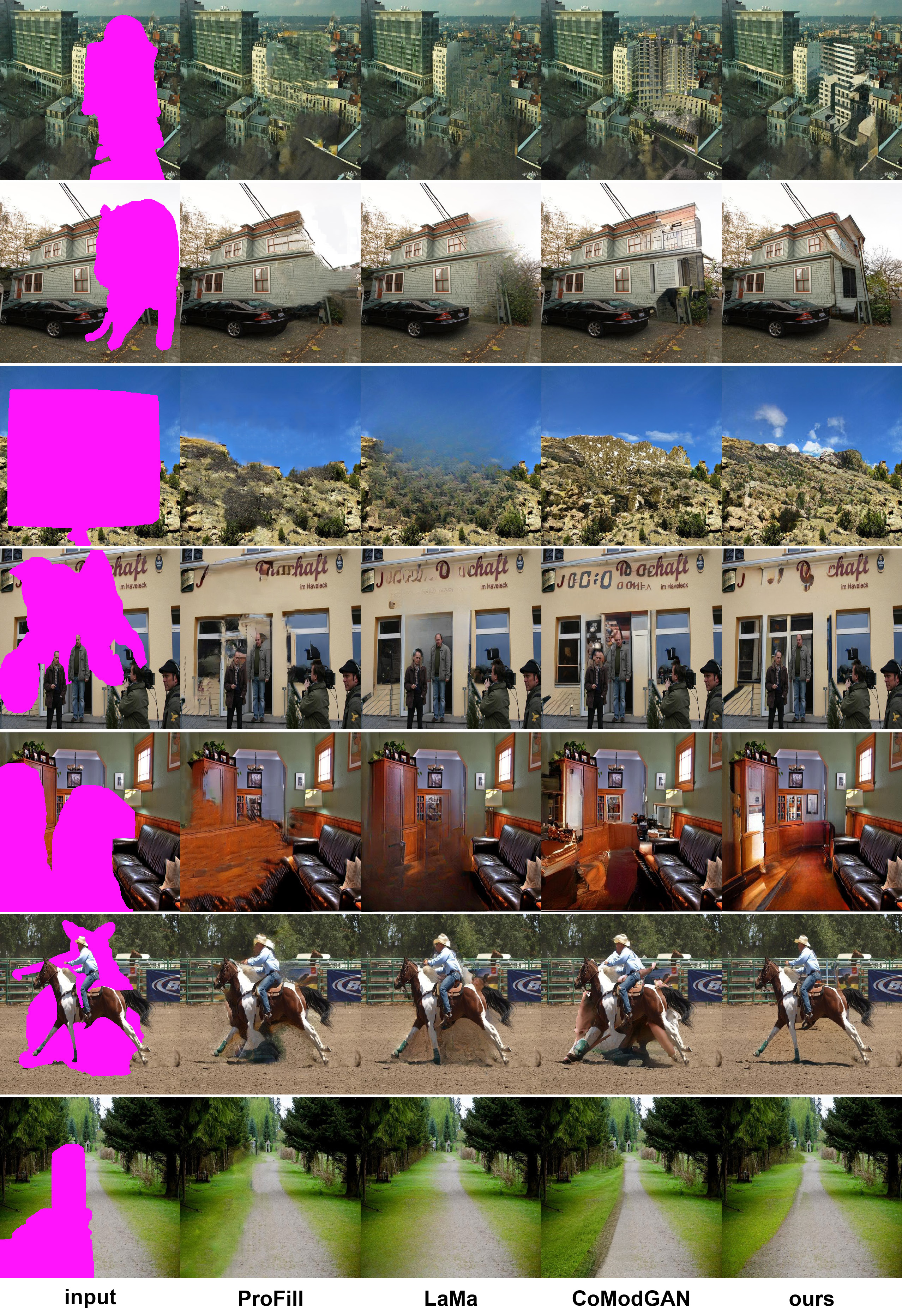}
	
	\caption{
		\small
		{
			\textbf{Results of CM-GAN in comparison to ProFill~\cite{profill}, LaMa~\cite{lama} and CoModGAN~\cite{comodgan}.}
			Best viewed by zoom-in on screen.
		}
	}
	\label{supp:fig:main-compare3}
\end{figure*}

\begin{figure*}[]
	\centering
	\includegraphics[width=1\linewidth]{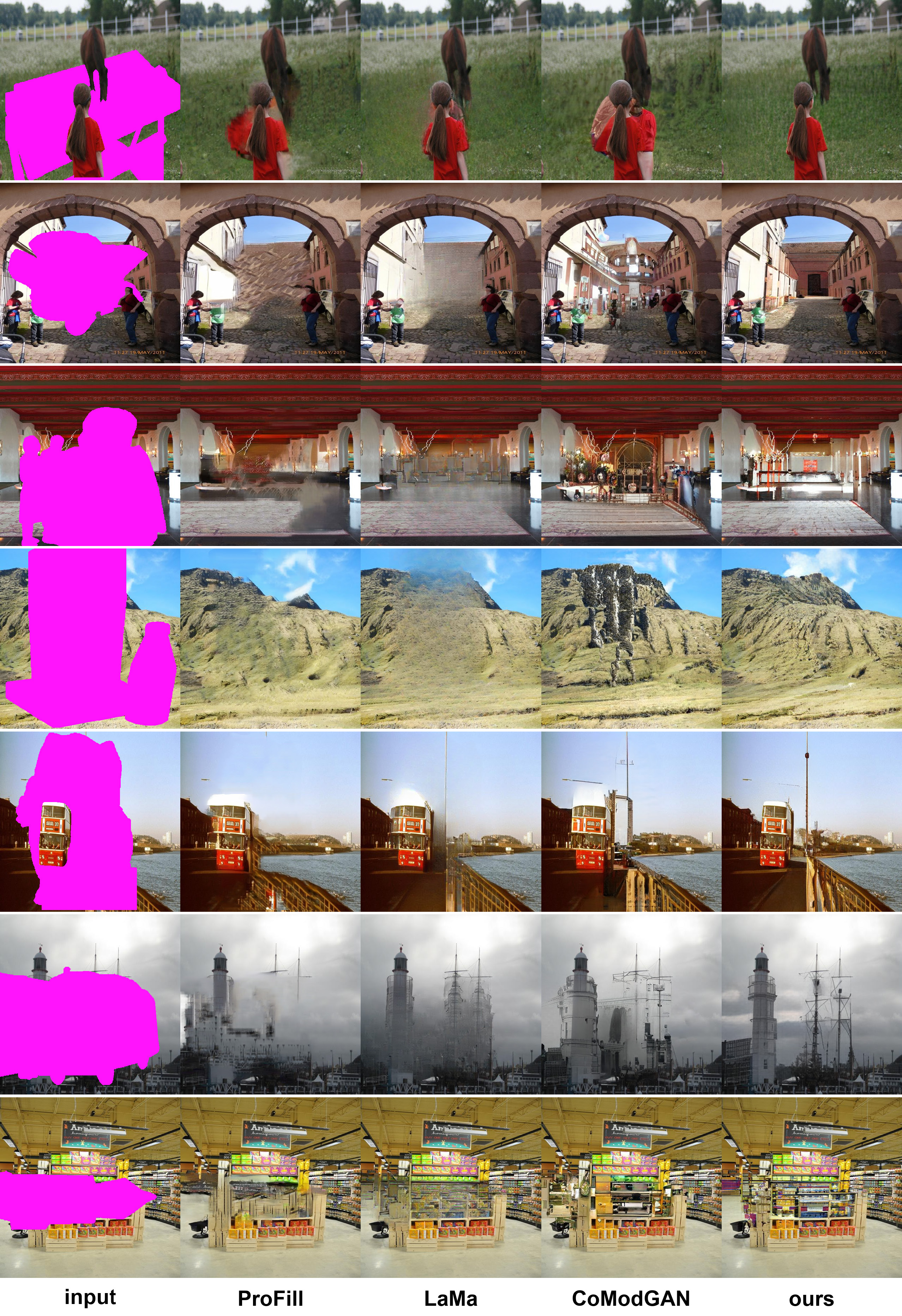}
	
	\caption{
		\small
		{
			\textbf{Results of CM-GAN in comparison to ProFill~\cite{profill}, LaMa~\cite{lama} and CoModGAN~\cite{comodgan}.}
			Best viewed by zoom-in on screen.
		}
	}
	\label{supp:fig:main-compare4}
\end{figure*}

We present additional visual comparisons to ProFill~\cite{profill}, LaMa~\cite{lama} and CoModGAN~\cite{comodgan} in \Cref{supp:fig:main-compare1,supp:fig:main-compare2,supp:fig:main-compare3,supp:fig:main-compare4} to supplement Figure 5 of the main paper.

\subsection{Visual Comparisons to Transformer-based Methods}
\label{supp:subsec:transformer}
\begin{figure*}[]
	\centering
	\includegraphics[width=1\linewidth]{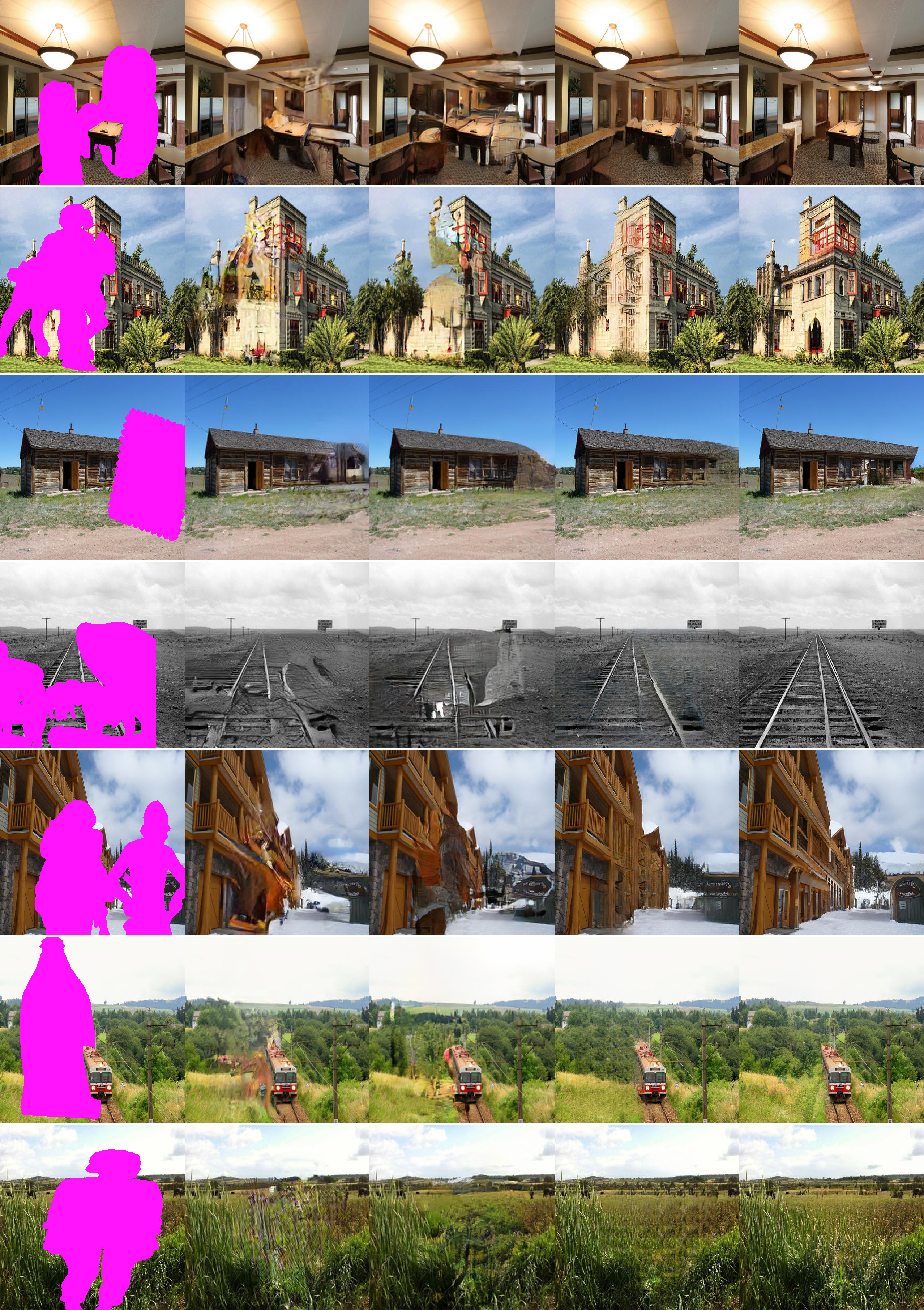}
	
	\includegraphics[width=1\linewidth]{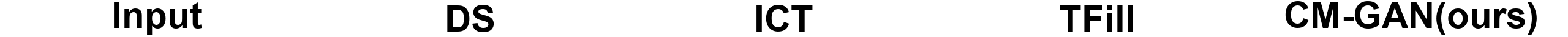}
	\caption{
		\small
		{
			\textbf{Results of CM-GAN in comparison to transformer-based approaches including DS~\cite{diverse}, ICT~\cite{ict} and TFill~\cite{zheng2021tfill}.}
			Best viewed by zoom-in on screen.
		}
	}
	\label{supp:fig:transformer-compare}
\end{figure*}

CM-GAN is based on the GAN framework. With transformers becoming popular in computer vision, several recent works~\cite{diverse,ict,zheng2021tfill} leverage transformer-based architectures for inpainting. In this section, we present the visual comparison to DS~\cite{diverse}, ICT~\cite{ict} and the recently proposed TFill~\cite{zheng2021tfill} to analyze the visual quality of those approaches. As observed in \Cref{supp:fig:transformer-compare}, CM-GAN achieves consistently better visual results in terms of holistic structures and local textures, which is coherent to the FID scores reported in Table 1 of the main paper \footnote{The FID score of TFill is 7.435.}.

\subsection{Visual Comparisons to Other Remaining Methods}
\label{supp:subsec:other}
\begin{figure*}[]
	\centering
	\includegraphics[width=1\linewidth]{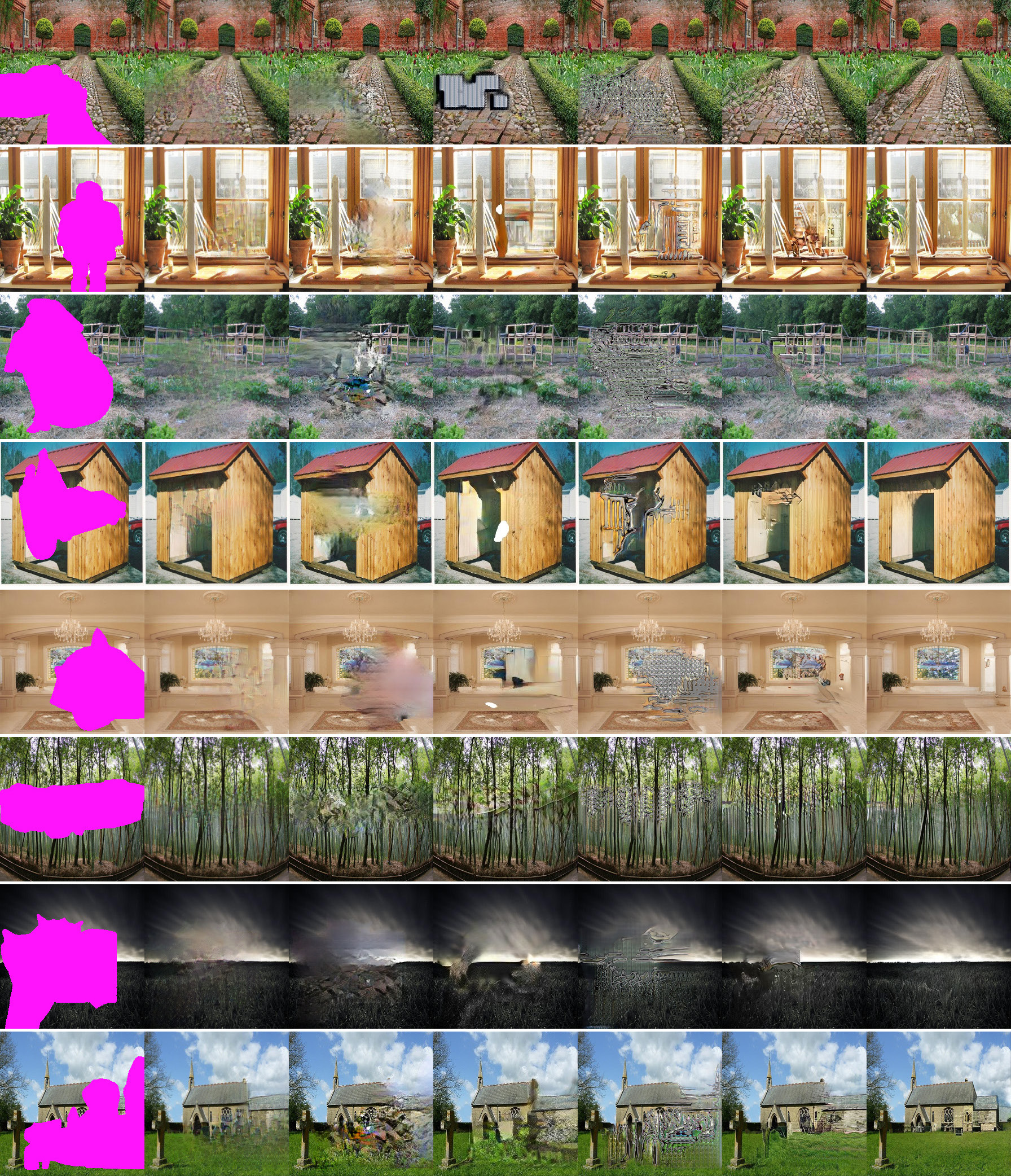}

	\includegraphics[width=1\linewidth]{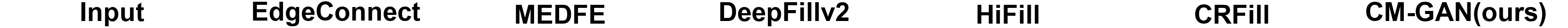}
	\caption{
		\small
		{
			\textbf{Results of CM-GAN in comparison to other methods including EdgeConnect~\cite{edgeconnect}, MEDEF~\cite{MEDFE}, DeepFillv2~\cite{deepfillv2}, HiFill~\cite{hifill}, CRFill~\cite{crfill}.
			}
			Best viewed by zoom-in on screen.
		}
	}
	\label{supp:fig:all-compare}
\end{figure*}

\Cref{supp:fig:all-compare} presents the visual comparisons of CM-GAN to other methods including EdgeConnect~\cite{edgeconnect}, MEDEF~\cite{MEDFE}, DeepFillv2~\cite{deepfillv2}, HiFill~\cite{hifill}, CRFill~\cite{crfill}. Our method achieves substantially better visual quality than all these compared methods.

\section{Visual Comparisons on Other Types of Masks}
\label{supp:sec:other-mask}
\begin{figure*}[]
	\centering
	\includegraphics[width=1\linewidth]{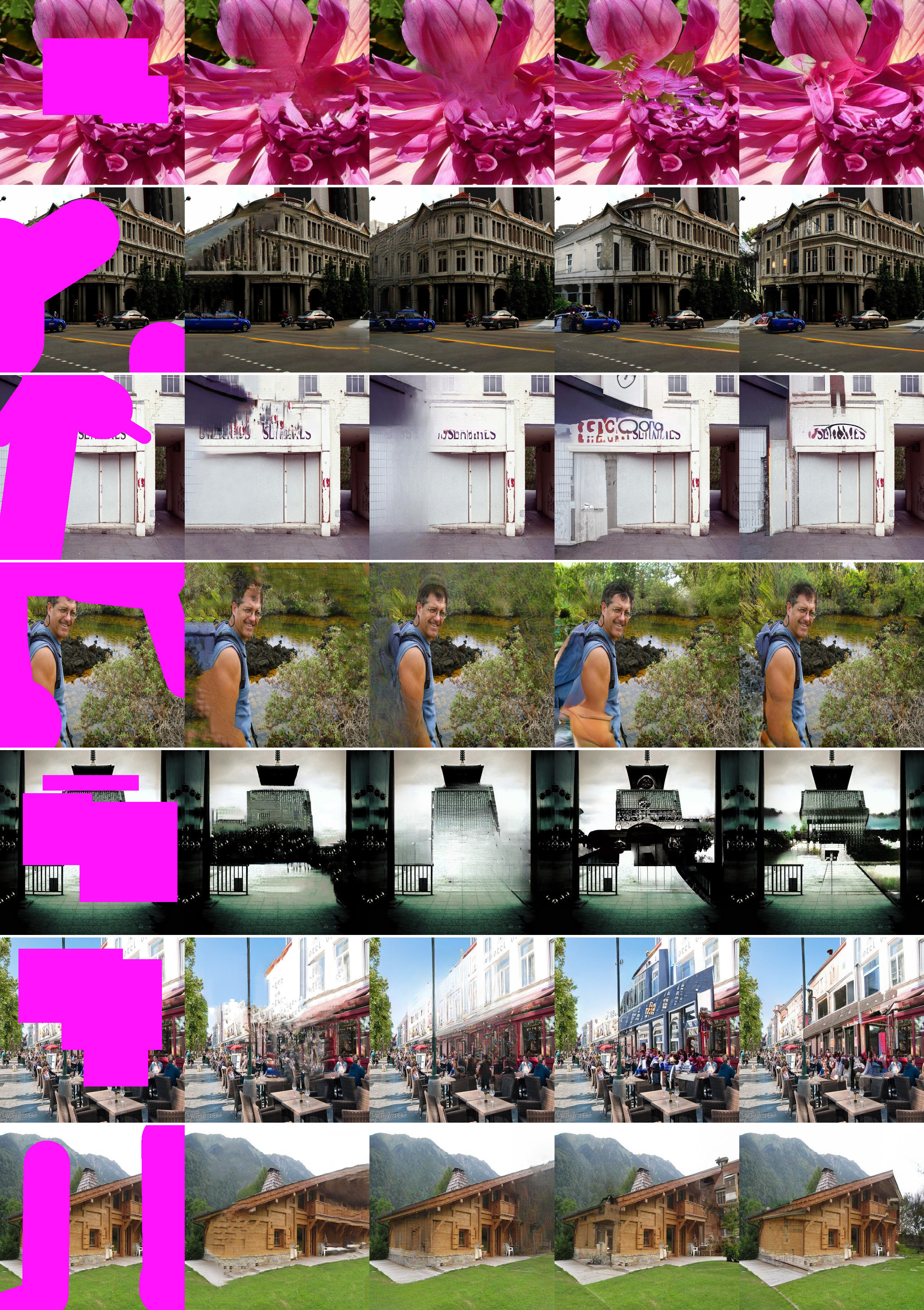}
	
	\includegraphics[width=1\linewidth]{figure_compressed/caption.png}
	\caption{
		\small
		{
			\textbf{Visual comparison on the mask of LaMa~\cite{lama}.
			}
			Best viewed by zoom-in on screen.
		}
	}
	\label{supp:fig:lama-mask}
\end{figure*}

To supplement Table 3 of the main paper, we provide the visual comparisons on the LaMa mask~\cite{lama} in \Cref{supp:fig:lama-mask} and CoModGAN mask~\cite{comodgan} in \Cref{supp:fig:comodgan-mask}, respectively.
Consistent to the numerical result in the main paper, our method generates better global structure and more coherent textures than others, demonstrating robustness of CM-GAN on different mask types.

\begin{figure*}[]
	\centering
	\includegraphics[width=1\linewidth]{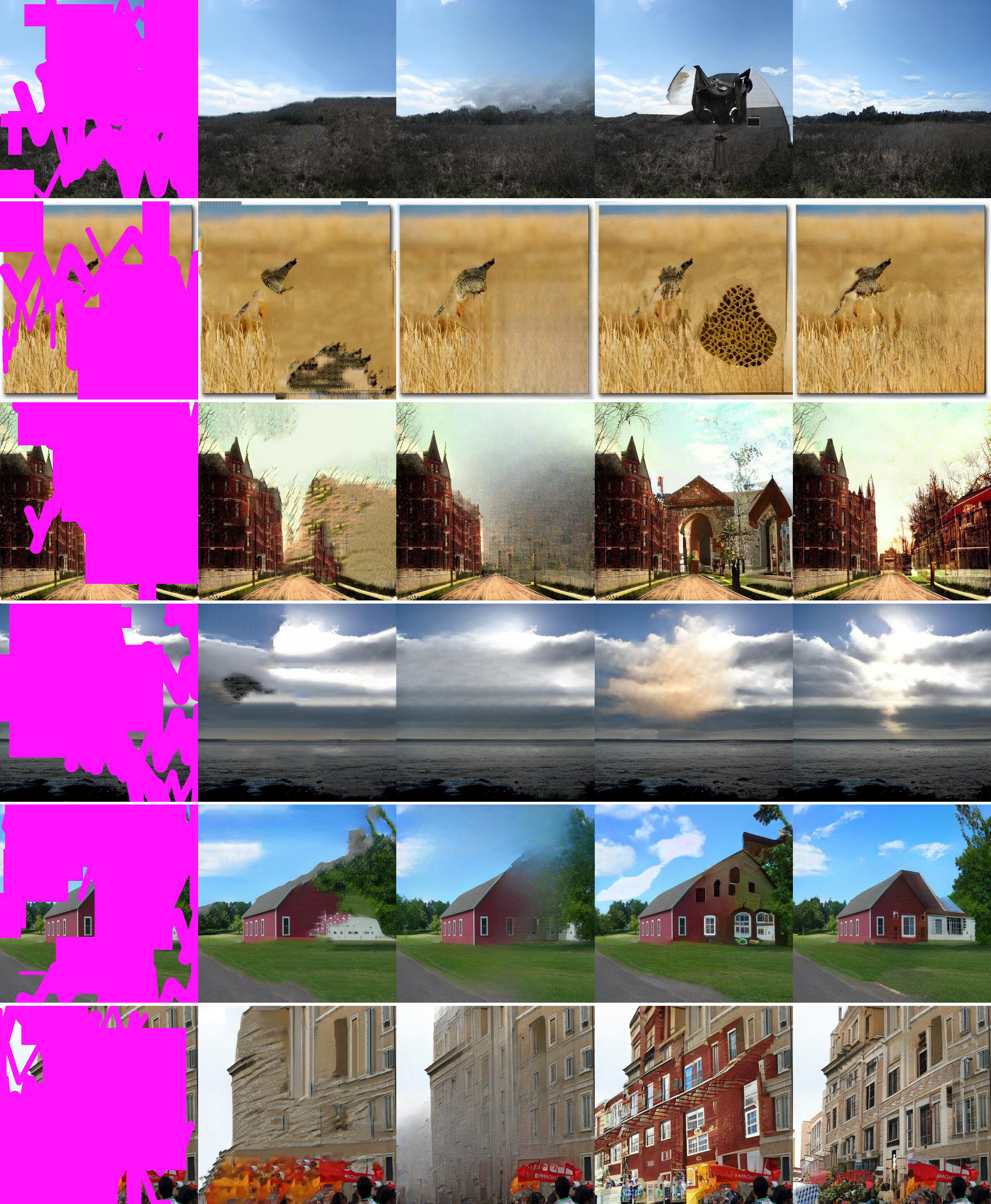}
	
	\includegraphics[width=1\linewidth]{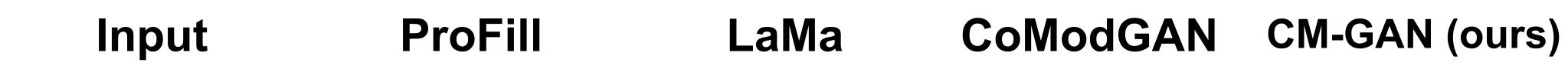}
	\caption{
		\small
		{
			\textbf{Visual comparison on the mask of CoModGAN~\cite{comodgan}. 
			}
			Best viewed by zoom-in on screen.
		}
	}
	\label{supp:fig:comodgan-mask}
\end{figure*}


\section{Implementation Details}
\label{supp:sec:implementation-details}

\subsection{The Details of Spatial Modulation}
We provide the detail implementation of the Affine Parameters Networks (APN) and our spatial modulation operation in pseudo code.

\noindent \textbf{The Affine Parameters Network (APN).} \quad
The affine parameters network (APN) is implemented as a stack of convolutional layer that takes $\tX$ as input to generate scaling parameters $\tA$ and shifting parameters $\tB$.
\begin{lstlisting}[language=Python]
def APN(X):
# the 1x1 input layer
t1 = self.conv1_1x1(X)
# the 3x3+1x1 middle layer
t2 = self.conv2_3x3(t1)
t2 = t2 + self.conv2_1x1(t1)
# the 1x1 output layer
A = self.conv_A_1x1(t)
B = self.conv_B_1x1(t)
return A, B
\end{lstlisting}

\noindent \textbf{Spatial Modulation.} \quad
Next, the spatial modulation takes feature maps $\tX$, $\tY$, global code $\vg$, the convolutional kernel weight $\vw$ and the noise $\vn$ as inputs to modulate $\tY$:
\begin{lstlisting}[language=Python]
import torch.nn.functional as F
def spatial_mod_ops(X, Y, g, w, noise):
bs = X.size(0) # batch size
# predicting the spatial code
A0, B = self.APN(X)
# merging with the global code
A = A0 + self.fc(g).reshape(bs,-1,1,1)
# spatial modulation
Y = Y.mul(A)
# convolution
Y = F.conv2d(Y, w)
# spatially-aware demodulation
w = w.unsqueeze(0)
A_avg_var = A.square().mean([2,3]).reshape(bs,1,-1,1,1)
D = (w.square().mul(A_avg_var).sum(dim=[2,3,4]) + 1e-8).rsqrt()
Y = Y.mul(D.reshape(bs, -1, 1, 1))
# adding bias and noise
Y = Y + B + noise
return Y
\end{lstlisting}


\subsection{Details of the Object-Aware Mask Generation Procedure}
Our object-aware mask generation scheme is based on the following pipeline to sample a mask for image $x$: 
\begin{enumerate}
    \item \emph{Generating the initial mask}. We sample an initial mask $\vm$ with either irregular masks proposed by ~\cite{comodgan}, object masks proposed by~\cite{profill} or random overlapping rectangle with probablity 0.45, 0.45 and 0.1, respectively. We further augment the object mask by random circular translation and dilate the mask using random width.
    \item \emph{Occluding foreground instance}. For each object instance $\vs_i$ from image $x$, we compute the overlapping ratio $r_i=\Area(\vm,\vs_i)/\Area(\vs_i)$ between the instance and the initial mask. If the overlapping ratio $r_i$ is larger than 0.5, we exclude instance $\vs_i$ from the initial mask $\vm$, namely $\vm \gets \vm - \vs_i$, to mimic the distractor removal use case.
    \item \emph{Rejecting small mask.} If the area of the mask is less than 0.05 of the area of the entire image, we repeating the sampling procedure, until the maximal sampling iteration 5 is reached. Otherwise, the sampled mask is returned.
\end{enumerate}


\clearpage
	\begin{thebibliography}{10}
\providecommand{\url}[1]{\texttt{#1}}
\providecommand{\urlprefix}{URL }
\providecommand{\doi}[1]{https://doi.org/#1}

\bibitem{albahar2021pose}
AlBahar, B., Lu, J., Yang, J., Shu, Z., Shechtman, E., Huang, J.B.: Pose with
  {S}tyle: {D}etail-preserving pose-guided image synthesis with conditional
  stylegan. ACM Transactions on Graphics  (2021)

\bibitem{aujol2006structure}
Aujol, J.F., Gilboa, G., Chan, T., Osher, S.: Structure-texture image
  decomposition—modeling, algorithms, and parameter selection. International
  journal of computer vision  \textbf{67}(1),  111--136 (2006)

\bibitem{ballester2001filling}
Ballester, C., Bertalmio, M., Caselles, V., Sapiro, G., Verdera, J.: Filling-in
  by joint interpolation of vector fields and gray levels. IEEE transactions on
  image processing  \textbf{10}(8),  1200--1211 (2001)

\bibitem{patchmatch}
Barnes, C., Shechtman, E., Finkelstein, A., Goldman, D.B.: Patchmatch: A
  randomized correspondence algorithm for structural image editing. ACM Trans.
  Graph.  \textbf{28}(3), ~24 (2009)

\bibitem{bertalmio2003simultaneous}
Bertalmio, M., Vese, L., Sapiro, G., Osher, S.: Simultaneous structure and
  texture image inpainting. IEEE transactions on image processing
  \textbf{12}(8),  882--889 (2003)

\bibitem{chan2001nontexture}
Chan, T.F., Shen, J.: Nontexture inpainting by curvature-driven diffusions.
  Journal of visual communication and image representation  \textbf{12}(4),
  436--449 (2001)

\bibitem{chen2019toward}
Chen, B.C., Kae, A.: Toward realistic image compositing with adversarial
  learning. In: Proceedings of the IEEE/CVF Conference on Computer Vision and
  Pattern Recognition. pp. 8415--8424 (2019)

\bibitem{ffc}
Chi, L., Jiang, B., Mu, Y.: Fast fourier convolution. Advances in Neural
  Information Processing Systems  \textbf{33} (2020)

\bibitem{cho2008patch}
Cho, T.S., Butman, M., Avidan, S., Freeman, W.T.: The patch transform and its
  applications to image editing. In: 2008 IEEE Conference on Computer Vision
  and Pattern Recognition. pp.~1--8. IEEE (2008)

\bibitem{criminisi2004region}
Criminisi, A., P{\'e}rez, P., Toyama, K.: Region filling and object removal by
  exemplar-based image inpainting. IEEE Transactions on image processing
  \textbf{13}(9),  1200--1212 (2004)

\bibitem{Image_quilting}
Efros, A.A., Freeman, W.T.: Image quilting for texture synthesis and transfer.
  In: Proceedings of the 28th annual conference on Computer graphics and
  interactive techniques. pp. 341--346. ACM (2001)

\bibitem{gan}
Goodfellow, I., Pouget-Abadie, J., Mirza, M., Xu, B., Warde-Farley, D., Ozair,
  S., Courville, A., Bengio, Y.: Generative adversarial nets. In: Advances in
  neural information processing systems. pp. 2672--2680 (2014)

\bibitem{gulrajani2017improved}
Gulrajani, I., Ahmed, F., Arjovsky, M., Dumoulin, V., Courville, A.: Improved
  training of wasserstein gans. arXiv preprint arXiv:1704.00028  (2017)

\bibitem{fid}
Heusel, M., Ramsauer, H., Unterthiner, T., Nessler, B., Hochreiter, S.: Gans
  trained by a two time-scale update rule converge to a local nash equilibrium.
  In: Advances in Neural Information Processing Systems. pp. 6626--6637 (2017)

\bibitem{MEDFE}
Hongyu~Liu, Bin~Jiang, Y.S.W.H., Chao, Y.: Rethinking image inpainting via a
  mutual encoder-decoder with feature equalizations. In: Proceedings of the
  European Conference on Computer Vision (2020)

\bibitem{huang2017arbitrary}
Huang, X., Belongie, S.: Arbitrary style transfer in real-time with adaptive
  instance normalization. In: Proceedings of the IEEE International Conference
  on Computer Vision. pp. 1501--1510 (2017)

\bibitem{huang2018multimodal}
Huang, X., Liu, M.Y., Belongie, S., Kautz, J.: Multimodal unsupervised
  image-to-image translation. In: Proceedings of the European conference on
  computer vision (ECCV). pp. 172--189 (2018)

\bibitem{iizuka2017globally}
Iizuka, S., Simo-Serra, E., Ishikawa, H.: Globally and locally consistent image
  completion. ACM Transactions on Graphics (ToG)  \textbf{36}(4),  1--14 (2017)

\bibitem{johnson2016perceptual}
Johnson, J., Alahi, A., Fei-Fei, L.: Perceptual losses for real-time style
  transfer and super-resolution. In: European conference on computer vision.
  pp. 694--711. Springer (2016)

\bibitem{stylegan}
Karras, T., Laine, S., Aila, T.: A style-based generator architecture for
  generative adversarial networks. In: Proceedings of the IEEE/CVF Conference
  on Computer Vision and Pattern Recognition. pp. 4401--4410 (2019)

\bibitem{stylegan2}
Karras, T., Laine, S., Aittala, M., Hellsten, J., Lehtinen, J., Aila, T.:
  Analyzing and improving the image quality of {StyleGAN}. In: Proc. CVPR
  (2020)

\bibitem{stylemapgan}
Kim, H., Choi, Y., Kim, J., Yoo, S., Uh, Y.: Exploiting spatial dimensions of
  latent in gan for real-time image editing. In: Proceedings of the IEEE
  Conference on Computer Vision and Pattern Recognition (2021)

\bibitem{adam}
Kingma, D.P., Ba, J.: Adam: A method for stochastic optimization. arXiv
  preprint arXiv:1412.6980  (2014)

\bibitem{Kopf-OneShot-2020}
Kopf, J., Matzen, K., Alsisan, S., Quigley, O., Ge, F., Chong, Y., Patterson,
  J., Frahm, J.M., Wu, S., Yu, M., Zhang, P., He, Z., Vajda, P., Saraf, A.,
  Cohen, M.: One shot 3d photography  \textbf{39}(4) (2020)

\bibitem{kwatra2005texture}
Kwatra, V., Essa, I., Bobick, A., Kwatra, N.: Texture optimization for
  example-based synthesis. In: ACM SIGGRAPH 2005 Papers, pp. 795--802 (2005)

\bibitem{panopticFCN}
Li, Y., Zhao, H., Qi, X., Wang, L., Li, Z., Sun, J., Jia, J.: Fully
  convolutional networks for panoptic segmentation. In: Proceedings of the
  IEEE/CVF Conference on Computer Vision and Pattern Recognition. pp. 214--223
  (2021)

\bibitem{partial_conv}
Liu, G., Reda, F.A., Shih, K.J., Wang, T.C., Tao, A., Catanzaro, B.: Image
  inpainting for irregular holes using partial convolutions. In: Proceedings of
  the European Conference on Computer Vision (ECCV). pp. 85--100 (2018)

\bibitem{luo2016understanding}
Luo, W., Li, Y., Urtasun, R., Zemel, R.: Understanding the effective receptive
  field in deep convolutional neural networks. In: Proceedings of the 30th
  International Conference on Neural Information Processing Systems. pp.
  4905--4913 (2016)

\bibitem{mescheder2018training}
Mescheder, L., Geiger, A., Nowozin, S.: Which training methods for gans do
  actually converge? In: International conference on machine learning. pp.
  3481--3490. PMLR (2018)

\bibitem{miyato2018spectral}
Miyato, T., Kataoka, T., Koyama, M., Yoshida, Y.: Spectral normalization for
  generative adversarial networks. arXiv preprint arXiv:1802.05957  (2018)

\bibitem{edgeconnect}
Nazeri, K., Ng, E., Joseph, T., Qureshi, F.Z., Ebrahimi, M.: Edgeconnect:
  Generative image inpainting with adversarial edge learning. arXiv preprint
  arXiv:1901.00212  (2019)

\bibitem{Niklaus_TOG_2019}
Niklaus, S., Mai, L., Yang, J., Liu, F.: 3d ken burns effect from a single
  image. ACM Transactions on Graphics  \textbf{38}(6),  184:1--184:15 (2019)

\bibitem{sesame}
Ntavelis, E., Romero, A., Kastanis, I., Gool, L.V., Timofte, R.: Sesame:
  Semantic editing of scenes by adding, manipulating or erasing objects. In:
  European Conference on Computer Vision. pp. 394--411. Springer (2020)

\bibitem{oh2001image}
Oh, B.M., Chen, M., Dorsey, J., Durand, F.: Image-based modeling and photo
  editing. In: Proceedings of the 28th annual conference on Computer graphics
  and interactive techniques. pp. 433--442 (2001)

\bibitem{spade}
Park, T., Liu, M.Y., Wang, T.C., Zhu, J.Y.: Semantic image synthesis with
  spatially-adaptive normalization. In: Proceedings of the IEEE Conference on
  Computer Vision and Pattern Recognition (2019)

\bibitem{park2020swapping}
Park, T., Zhu, J.Y., Wang, O., Lu, J., Shechtman, E., Efros, A.A., Zhang, R.:
  Swapping autoencoder for deep image manipulation. arXiv preprint
  arXiv:2007.00653  (2020)

\bibitem{pathak2016context}
Pathak, D., Krahenbuhl, P., Donahue, J., Darrell, T., Efros, A.A.: Context
  encoders: Feature learning by inpainting. In: Proceedings of the IEEE
  conference on computer vision and pattern recognition. pp. 2536--2544 (2016)

\bibitem{diverse}
Peng, J., Liu, D., Xu, S., Li, H.: Generating diverse structure for image
  inpainting with hierarchical vq-vae. In: Proceedings of the IEEE/CVF
  Conference on Computer Vision and Pattern Recognition (CVPR). pp.
  10775--10784 (2021)

\bibitem{structureflow}
Ren, Y., Yu, X., Zhang, R., Li, T.H., Liu, S., Li, G.: Structureflow: Image
  inpainting via structure-aware appearance flow. In: IEEE International
  Conference on Computer Vision (ICCV) (2019)

\bibitem{salimans2016weight}
Salimans, T., Kingma, D.P.: Weight normalization: A simple reparameterization
  to accelerate training of deep neural networks. Advances in neural
  information processing systems  \textbf{29},  901--909 (2016)

\bibitem{setlur2005automatic}
Setlur, V., Takagi, S., Raskar, R., Gleicher, M., Gooch, B.: Automatic image
  retargeting. In: Proceedings of the 4th international conference on Mobile
  and ubiquitous multimedia. pp. 59--68 (2005)

\bibitem{shen2002mathematical}
Shen, J., Chan, T.F.: Mathematical models for local nontexture inpaintings.
  SIAM Journal on Applied Mathematics  \textbf{62}(3),  1019--1043 (2002)

\bibitem{song2018spg}
Song, Y., Yang, C., Shen, Y., Wang, P., Huang, Q., Kuo, C.C.J.: Spg-net:
  Segmentation prediction and guidance network for image inpainting. arXiv
  preprint arXiv:1805.03356  (2018)

\bibitem{lama}
Suvorov, R., Logacheva, E., Mashikhin, A., Remizova, A., Ashukha, A.,
  Silvestrov, A., Kong, N., Goka, H., Park, K., Lempitsky, V.:
  Resolution-robust large mask inpainting with fourier convolutions. arXiv
  preprint arXiv:2109.07161  (2021)

\bibitem{tan2020semantic}
Tan, Z., Chen, D., Chu, Q., Chai, M., Liao, J., He, M., Yuan, L., Hua, G., Yu,
  N.: Semantic image synthesis via efficient class-adaptive normalization.
  arXiv preprint arXiv:2012.04644  (2020)

\bibitem{vaquero2010survey}
Vaquero, D., Turk, M., Pulli, K., Tico, M., Gelfand, N.: A survey of image
  retargeting techniques. In: Applications of Digital Image Processing XXXIII.
  vol.~7798, pp. 328--342. SPIE (2010)

\bibitem{ict}
Wan, Z., Zhang, J., Chen, D., Liao, J.: High-fidelity pluralistic image
  completion with transformers. arXiv preprint arXiv:2103.14031  (2021)

\bibitem{wang2018recovering}
Wang, X., Yu, K., Dong, C., Loy, C.C.: Recovering realistic texture in image
  super-resolution by deep spatial feature transform. In: Proceedings of the
  IEEE conference on computer vision and pattern recognition. pp. 606--615
  (2018)

\bibitem{foreground_aware}
Xiong, W., Yu, J., Lin, Z., Yang, J., Lu, X., Barnes, C., Luo, J.:
  Foreground-aware image inpainting. In: Proceedings of the IEEE/CVF Conference
  on Computer Vision and Pattern Recognition. pp. 5840--5848 (2019)

\bibitem{YangLLSWL17}
Yang, C., Lu, X., Lin, Z., Shechtman, E., Wang, O., Li, H.: High-resolution
  image inpainting using multi-scale neural patch synthesis. In: 2017 {IEEE}
  Conference on Computer Vision and Pattern Recognition, {CVPR} 2017, Honolulu,
  HI, USA, July 21-26, 2017. pp. 4076--4084 (2017)

\bibitem{yang2020learning}
Yang, J., Qi, Z., Shi, Y.: Learning to incorporate structure knowledge for
  image inpainting. In: Proceedings of the AAAI Conference on Artificial
  Intelligence. vol.~34, pp. 12605--12612 (2020)

\bibitem{hifill}
Yi, Z., Tang, Q., Azizi, S., Jang, D., Xu, Z.: Contextual residual aggregation
  for ultra high-resolution image inpainting. In: Proceedings of the IEEE/CVF
  Conference on Computer Vision and Pattern Recognition. pp. 7508--7517 (2020)

\bibitem{yu2015multi}
Yu, F., Koltun, V.: Multi-scale context aggregation by dilated convolutions.
  arXiv preprint arXiv:1511.07122  (2015)

\bibitem{deepfill}
Yu, J., Lin, Z., Yang, J., Shen, X., Lu, X., Huang, T.S.: Generative image
  inpainting with contextual attention. In: Proceedings of the IEEE conference
  on computer vision and pattern recognition. pp. 5505--5514 (2018)

\bibitem{deepfillv2}
Yu, J., Lin, Z., Yang, J., Shen, X., Lu, X., Huang, T.S.: Free-form image
  inpainting with gated convolution. In: Proceedings of the IEEE International
  Conference on Computer Vision. pp. 4471--4480 (2019)

\bibitem{crfill}
Zeng, Y., Lin, Z., Lu, H., Patel, V.M.: Cr-fill: Generative image inpainting
  with auxiliary contextual reconstruction. In: Proceedings of the IEEE
  International Conference on Computer Vision (2021)

\bibitem{profill}
Zeng, Y., Lin, Z., Yang, J., Zhang, J., Shechtman, E., Lu, H.: High-resolution
  image inpainting with iterative confidence feedback and guided upsampling.
  arXiv preprint arXiv:2005.11742  (2020)

\bibitem{lpips}
Zhang, R., Isola, P., Efros, A.A., Shechtman, E., Wang, O.: The unreasonable
  effectiveness of deep features as a perceptual metric. In: Proceedings of the
  IEEE Conference on Computer Vision and Pattern Recognition. pp. 586--595
  (2018)

\bibitem{comodgan}
Zhao, S., Cui, J., Sheng, Y., Dong, Y., Liang, X., Chang, E.I., Xu, Y.: Large
  scale image completion via co-modulated generative adversarial networks.
  arXiv preprint arXiv:2103.10428  (2021)

\bibitem{zheng2021tfill}
Zheng, C., Cham, T.J., Cai, J.: Tfill: Image completion via a transformer-based
  architecture. arXiv preprint arXiv:2104.00845  (2021)

\bibitem{zheng2020example}
Zheng, H., Liao, H., Chen, L., Xiong, W., Chen, T., Luo, J.: Example-guided
  image synthesis using masked spatial-channel attention and self-supervision.
  In: European Conference on Computer Vision. pp. 422--439. Springer (2020)

\bibitem{zhou2017places}
Zhou, B., Lapedriza, A., Khosla, A., Oliva, A., Torralba, A.: Places: A 10
  million image database for scene recognition. IEEE transactions on pattern
  analysis and machine intelligence  \textbf{40}(6),  1452--1464 (2017)

\end{thebibliography}
\end{document}